\documentclass[10pt,twocolumn,letterpaper]{article}

\usepackage{times}
\usepackage{epsfig}
\usepackage{graphicx}
\usepackage{amsmath}
\usepackage{amssymb}
\usepackage{epstopdf}
\usepackage{subfigure}
\usepackage{color}
\usepackage{booktabs}
\usepackage{makecell}
\usepackage{geometry}

\begin{document}

%%%%%%%%% TITLE
\title{SID-NISM: A Self-supervised Low-light Image Enhancement Framework}

\author{Lijun Zhang\qquad Xiao Liu \qquad Erik Learned-Miller \qquad Hui Guan\\University of Massachusetts Amherst\\{\tt\small \{lijunzhang, xiaoliu1990, elm, huiguan\}@cs.umass.edu}}
\date{}

\maketitle
%\thispagestyle{empty}

%%%%%%%%% ABSTRACT
\begin{abstract}
   When capturing images in low-light conditions, the images often suffer from low visibility, which not only degrades the visual aesthetics of images, but also significantly degenerates the performance of many computer vision algorithms. In this paper, we propose a self-supervised low-light image enhancement framework (SID-NISM), which consists of two components, a Self-supervised Image Decomposition Network (SID-Net) and a Nonlinear Illumination Saturation Mapping function (NISM). As a self-supervised network, SID-Net could decompose the given low-light image into its reflectance, illumination and noise directly without any prior training or reference image, which distinguishes it from existing supervised-learning methods greatly. Then, the decomposed illumination map will be enhanced by NISM. Having the restored illumination map, the enhancement can be achieved accordingly. Experiments on several public challenging low-light image datasets reveal that the images enhanced by SID-NISM are more natural and have less unexpected artifacts.
\end{abstract}

\section{Introduction}\label{sect:intro}
Generally, capturing high-quality images in dim light or back light conditions is challenging, since insufficient lighting can significantly degrade the visibility of images. Especially the lost details and low contrast would not only cause unpleasant subjective perceptions, but also hurt the performance of back-end computer vision systems which are designed for normal-light images. Though modern imaging sensors can automatically set high ISO, long exposure, and flash according to different circumstances to compensate for the low light, they suffer from different drawbacks.
% For instance, high ISO increases the sensitivity of an image sensor to light, but the noise is also amplified.
One solution is to use high dynamic range (HDR) imaging techniques, which has been integrated into modern cameras to tackle with dark light environment in the image acquisition stage. However, when it comes to restore existing poor-quality low-light images, HDR needs bunch of images under different illumination conditions as inputs, which limits its practical application.

Thus the low-light image enhancement has been a long-standing problem in the community with a great progress made over the past years.
It can not only be used to increase the visual aesthetics of photos for people's daily use, but also to stable the performance of many computer vision algorithms such as object detection. Existing methods have two mainstream ideologies. One tries to consider the low-light enhancement problem as an image decomposition problem based on the Retinex theory, in which each image can be separated into independent components and the image reflectance part can be seen as a reasonable enhancement result \cite{fu2016weighted,guo2016lime,ren2018joint,zhang2018high}. However, these methods tend to introduce unexpected artifacts in the enhanced results. The other resorts to machine learning techniques based on large-scale image databases consisting of pairs of low-light image and corresponding enhanced image restored by image processing software like Photoshop \cite{wei2018deep,dale2009image,bychkovsky2011learning,shen2017msr}. These methods have achieved impressive performance but usually limited by the quality and quantity of the training database.

To get rid of the restriction of using training data, in this paper, we present a novel self-supervised low-light image enhancement framework called SID-NISM, which could restore the quality of any single low-light image only relying on the visual information of the image itself. Our major contributions are summarized as follows.

(1) A self-supervised image decomposition network, SID-Net, is proposed to decompose the input image into lighting-independent reflectance, structure-aware smooth illumination and reflectance-related noise straightforwardly according to the robust Retinex theory~\cite{li2018structure}. Taking as inputs any given input image and its corresponding histogram equalization image, SID-Net can converge to the optimal decomposed maps within limited iterations, in which two novel loss terms related to the formation of the reflectance and noise maps together with several common-used basic terms are adopted to guide the decomposition procedure preciously. As pointed out above, SID-Net is a image-specific network without depending on any prior training or reference image, which greatly distinguish it from existing supervised-learning methods like Retinex-Net~\cite{wei2018deep}.

(2) A nonlinear illumination saturation mapping function (NISM) is constructed to refine the decomposed low-light illumination map. Specifically, it could brighten up the whole image to a proper lighting level on the premise of preserving the contrast between foreground and background, which combats the weaknesses of Gamma correction in contrast preservation and bright regions enhancement. The final normal-light result can be restored by combining the denoised reflectance with the enhanced illumination.

Comprehensive experiments are conducted to illustrate the performance of the proposed method. Human perception user study suggests that people are more inclined to prefer the output of our method and find less unexpected artifacts in our results when the results of multiple competing methods are presented in front of them, which is consistent with the objective evaluations. Both subjective and objective experiments demonstrate the superiority of our method over the state-of-the-art methods in producing natural and attractive enhancement results.

\section{Related Work}\label{sect:related}
Researchers have devoted their efforts to solving the problem of low-light image enhancement in the past decades. Many techniques have been developed to improve the quality of low-light images, which can be classified into two major categories as mentioned in Sect. \ref{sect:intro}: Retinex-based methods and learning-based methods.

\textbf{Retinex-based Methods.} According to the classic Retinex theory~\cite{land1971lightness}, an observed image can be decomposed into two components, reflectance and illumination, in which the former one can be taken as a kind of compelling low-light image enhancement result. Jobson \textit{et al.}~\cite{jobson1997multiscale,jobson1997properties} made some early attempts based on this model, but their results are usually unrealistic. Wang \textit{et al.}~\cite{wang2013naturalness} presented a naturalness preserved method by utilizing a lightness-order-error measure. However, it tends to produce results with dim artifacts and requires expensive computational cost. Fu \textit{et al.}~\cite{fu2016weighted} and Ren \textit{et al.}~\cite{ren2018joint} succeed to simultaneously estimate reflectance and illumination maps through a weighted variational model and a sequential model respectively, but both suffered from preserving the property of Retinex model insufficiently. While Guo \textit{et al.}~\cite{guo2016lime} tired to only estimate the illumination map from the maximum values of three channels with constraints on preserving the main contour, then compute the reflectance by conducting element-wise division. And following the illumination estimation constraints in \cite{guo2016lime}, Zhang \textit{et al.}~\cite{zhang2018high} proposed two more constraints for estimating illumination based on perceptually bidirectional similarity. These two methods have pretty good performance but usually introduce unexpected artifacts in the enhanced images. Besides, with the development of neural network, some researchers~\cite{zhang2019kindling,wang2019underexposed,wei2018deep} sought to design image decomposition networks and illumination adjustment networks based on low-light and normal-light image pairs.

\textbf{Learning-based Methods.} Using machine learning tools to solve the problem of low-light image enhancement is a recent trend and also a promising direction. Dale \textit{et al.}~\cite{dale2009image} first established a database comprising 1 million images. Given an input image to be enhanced, their system executes a visual search to find the closest images in the database; these images define the input’s visual context, which can be further exploited to instantiate the restoration operations. Kang \textit{et al.}~\cite{kang2010personalization} constructed a database which stored the feature vectors describing training images along with vectors of enhancement parameters. Given a test image, the database was then searched for the best matching image, and the corresponding enhancement parameters were used to perform adjustment. Following the similar idea, Bychkovsky \textit{et al.}~\cite{bychkovsky2011learning} constructed a dataset containing 5,000 input-output image pairs that could be used to learn global tonal adjustments. To avoid collecting large-scale datasets, Shen \textit{et al.}~\cite{shen2017msr} trained their MSR-net designed upon the multi-scale Retinex theory on synthesized pairwise images.

With the power of neural network, these methods can get outstanding enhanced results, but at the same time it should be noticed that they are all based on supervised-learning frameworks and thus their performance highly depends on the quality and quantity of training datasets. Therefore Zhang \textit{et al.}~\cite{zhang2019zero} proposed the first unsupervised-learning back-lit image restoration network based on the S-curve theory~\cite{yuan2012automatic}, which estimated the image specific S-curve through region-level optimal exposure evaluation. However, it is limited by the mid-gray assumption \cite{guo2016lime} in S-curve, which leads to unrealistic enhanced results with halos and gray shadows.

\section{Retinex Theory}\label{sect:retinex}
The classic Retinex theory~\cite{land1971lightness} models the image compositions which could reflect the formation of low-light images to some extent. It assumes that the captured image can be decomposed into two components, reflectance and illumination. Let $S$ represents the source image, then the classic Retinex theory can be denoted by,
\begin{equation}\label{equ:retinex}
    S = R \times L
\end{equation}
where $R$ is reflectance, $L$ is illumination and $\times$ is element-wise multiplication. Illumination $L$ refers to the various lightness on observed objects, while reflectance $R$ corresponds to the material RGB color that describes how objects reflect light, which is considered to be invariant to $L$ and other possible imaging conditions. In short, $R$ represents the intrinsic property of captured objects. Besides, in color images each channel of $R$ can be regarded as sharing the same grayscale illumination map $L$.

Since low-light images usually suffer from darkness and unbalanced illumination distributions, the low-light image enhancement problem can be regarded as a procedure of estimating the illumination-independent reflectance according to the Retinex theory, which aims to remove the illumination effect and recover the original appearance of the scene objects. However, directly decomposing an input image into reflectance and illumination yields the intrinsic image decomposition problem~\cite{grosse2009ground}, which is inherently ill-posed and may produce unrealistic results~\cite{guo2016lime}. Therefore, some researchers~\cite{guo2016lime,fu2016weighted,ren2018joint,wei2018deep} converted the decomposition problem into an optimization problem and then computed the optimal solution through conventional optimization solvers or machine learning techniques. Moreover, instead of taking image reflectance as the enhancement result directly, existing methods usually project the enhanced illumination back to the reflectance by $R \times f(L)$ at the end, where $f(\cdot)$ stands for a manipulation operator adopted to enhance the illumination such as Gamma correction.

Another noticeable issue in low-light image enhancement is the unexpected noise raised by enhancing dark regions. Although the low-light source image doesn't suffer from the noise problem, the noise hidden in the dark would be amplified along with stretching the contrast of dark regions. To address the intensive noise problem, researchers seek for help from denoising algorithms like BM3D~\cite{dabov2007image} as the post-processing method, which is straightforward but not designed to solve the specific denoising problem here. Therefore we decide to follow the robust Retinex model~\cite{li2018structure}, which integrate the noise term $N$ into the classic Retinex theory directly,
\begin{equation}\label{equ:retinex}
    S = R \times L + N
\end{equation}
Once the source image could be decomposed into the three components successfully, the reflectance map would be not only independent with the illumination map but also get rid of unexpected noise.

\section{SID-NISM: Self-supervised Low-light Image Enhancement Framework} \label{sect:sid-nism}
\begin{figure*}[t]
    \centering
    \includegraphics[scale=0.32]{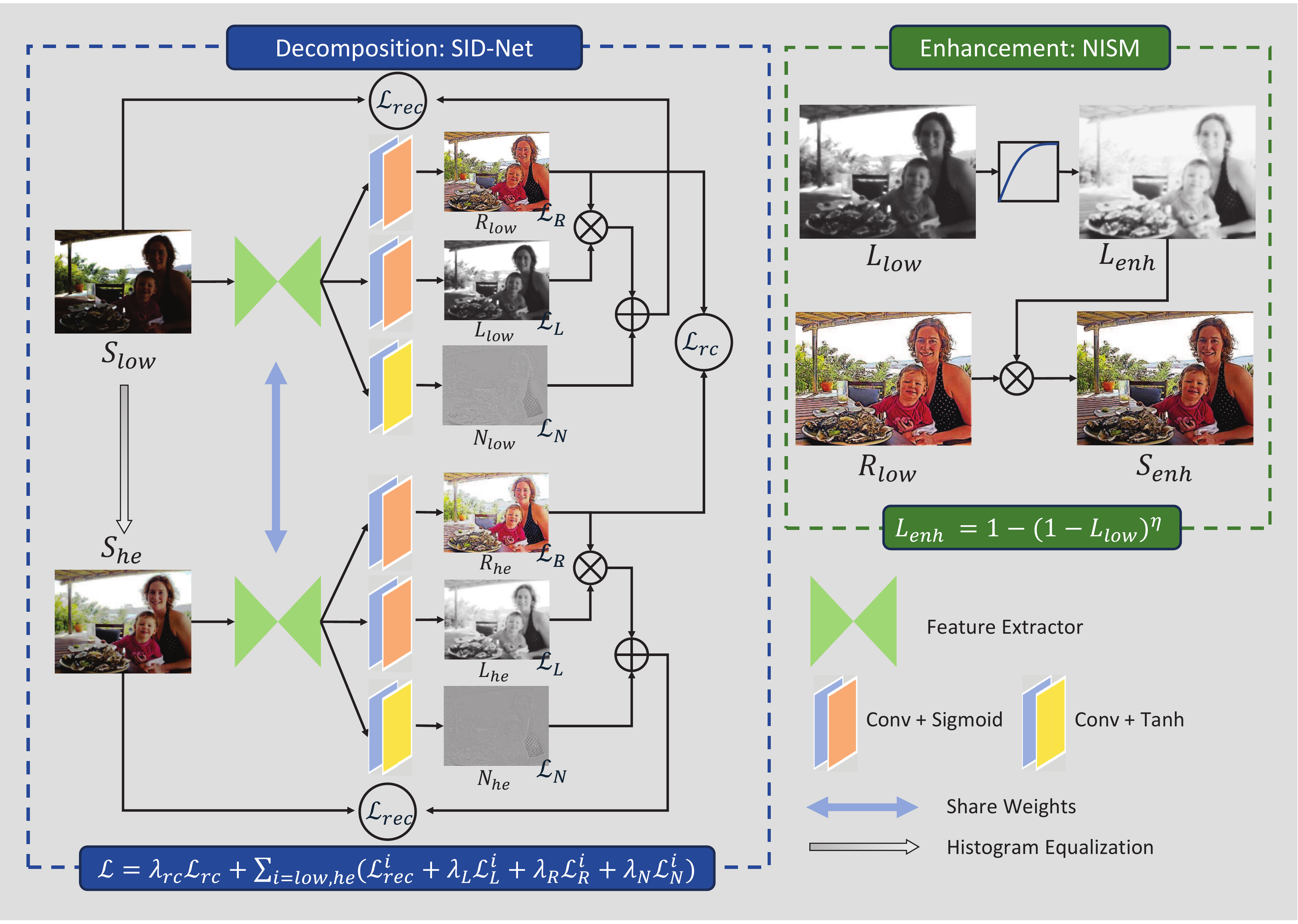}
    \caption{The overall structure of SID-NISM. It consists of two stages, decomposition and enhancement, which are corresponding to the two main parts SID-Net and NISM.}
    \label{fig:Framework}
\end{figure*}

In this section, the proposed low-light image enhancement framework SID-NISM is presented in details. Fig.~\ref{fig:Framework} illustrates the overall structure of SID-NISM, which consists of two stages, decomposition and enhancement, corresponding to the two main parts, SID-Net and NISM.

\subsection{SID-Net: A Self-supervised Image Decomposition Network}
According to the definition of the Retinex theory, the reflectance component of an image should be invariant to the illumination component and other imaging conditions, which suggests that any pair of images with the same content should share the same reflectance regardless of their different illumination conditions.
Thus inspired by the observation, a self-supervised learning image decomposition network called SID-Net is designed to split given low-light image into its reflectance, illumination and noise directly by taking as inputs the original image and its histogram equalization version. Notice that since the histogram equalization image is served as one of the paired images under different illumination condition to share the same reflectance, not the ground-truth normal-light images as in the supervised-learning methods, it doesn't matter whether histogram equalization would produce images with unpleasant color distortion or unbalanced illumination. Besides, although theoretically capturing paired images or utilizing other image generation methods are acceptable as well, considering the practical application and the computational cost, histogram equalization image becomes the best option.

As illustrated in Fig.~\ref{fig:Framework}, SID-Net firstly takes as inputs the source image $S_{low}$ and its corresponding histogram equalization version $S_{he}$ to split both of them into three components according to the robust Retinex theory, then iteratively converges to the optimal decomposition maps by minimizing the designed loss function.
Since it is a self-supervised network without external information from training images, the key point of SID-Net is to devise reasonable and effective loss function to guide the network to generate expected decomposition maps. The loss function of SID-Net is formulated as follows,
\begin{equation}
    \mathcal{L} = \lambda_{rc}\mathcal{L}_{rc} + \sum_{i=low,he} {(\mathcal{L}_{rec}^i + \lambda_{L}\mathcal{L}_{L}^i + \lambda_{R}\mathcal{L}_{R}^i + \lambda_{N}\mathcal{L}_{N}^i)}
\end{equation}
where $\lambda_{rc}$, $\lambda_{L}$, $\lambda_{R}$, $\lambda_{N}$ denote the coefficients to balance the reflectance similarities and the guidance of generating illumination, reflectance and noise maps. They are signi ﬁcantly smaller than 1 to address the importance of the fidelity term $\mathcal{L}_{rec}$ in the optimization, which are 0.01, 0.1, 0.001, 0.01 respectively in experiments. Notice that the reconstruction loss $\mathcal{L}_{rec}$ and the three guidance loss terms $\mathcal{L}_{L}, \mathcal{L}_{R}, \mathcal{L}_{N}$ are both applied on the original low-light image $S_{low}$ and the histogram equalization image $S_{he}$ to separate them into reflectance $R_{low}, R_{he}$, illumination $L_{low}, L_{he}$ and noise $N_{low}, N_{he}$ in a self-supervised way.

\textbf{Retinex reconstruction loss.} First and foremost, according to the robust Retinex theory~\cite{li2018structure}, the three decomposed maps should be able to reconstruct the original image. Thus the reconstruction loss $\mathcal{L}_{rec}$ is formulated as,
\begin{equation}
    \mathcal{L}_{rec} = \left\|R\times L+N-S\right\|_1
\end{equation}
It corresponds to the image formation, which could constrain the distance between estimated $R\times L+N$ and the source image $S$.

\textbf{Reflectance consistency loss.} As described in Sect.~\ref{sect:retinex}, the reflectance reflects the intrinsic property of captured objects, which should be invariant to the scene lighting and imaging conditions. That is to say, in the case of the same captured objects, i.e. image content, the reflectance maps of the low-light image and its histogram equalization version should be as close as possible. Therefore, based on the above assumption, the reflectance consistency loss $\mathcal{L}_{rc}$ is designed to constrain the reflectance similarities between $R_{low}$ and $R_{he}$.
\begin{equation}
    \mathcal{L}_{rc} = \left\|R_{low}-R_{he}\right\|_1
\end{equation}

\textbf{Illumination smoothness and consistency loss.} The loss function designed for the illumination map $\mathcal{L}_{L}$ is consist of two parts, the smoothness term and the consistency term, which can be expressed as,
\begin{equation}
\begin{split}
    \mathcal{L}_{L} &= {\left\|\nabla L \times \exp(-\alpha\nabla R)\right\|_1} \\
    &+ \left\|\nabla L \times \exp(-\alpha\sum_{j=low,he}{\nabla L_j})\right\|_1
\end{split}
\end{equation}
where $\nabla$ denotes the gradient including $\nabla_h$ (horizontal) and $\nabla_v$ (vertical) and $\alpha$ denotes the coefficient balancing the strength of structure-awareness and edge-preservation which is set as 10 in experiments.

The first term is adopted to guide the illumination map to be piece-wise smooth in textural details while preserving the general structure of the original images. A common-used smoothness prior in various image restoration tasks, total variation minimization (TV)~\cite{ma2012tv,ng2011total}, which minimizes the gradient of the whole image, is selected here as the fundamental of the illumination smoothness term. The second one is designed to preserve strong mutual edges while depressing weak ones. Similar to the illumination smoothness loss, it utilizes the weighted version of TV loss. Notice that since TV loss is structure-blindness, the original TV function is weighted with the gradient of the reflectance~\cite{wei2018deep} and illumination maps~\cite{zhang2019kindling} respectively to make the constructed loss be aware of the image structure and the two illumination maps be consistent with each other in terms of the image edges.

\textbf{Reflectance contrast and color loss.} To generate accurate reflectance map, in addition to the basic reconstruction loss, a novel reflectance contrast and color loss $\mathcal{L}_{R}$ is introduced to improve the image contrast and restore the image color as much as possible, which can be expressed as,
\begin{equation}
\begin{split}
    \mathcal{L}_{R} &= \frac{1}{3}\sum_{ch=R,G,B}{\left\|\nabla R^{ch} - \beta\nabla \hat{S}^{ch} \right\|_F} \\
    &+ \left\|R^H-S^H\right\|_2
\end{split}
\end{equation}
where $\beta$ is the gradient amplification factor which is set as 10 in experiments. $R^H$ and $S^H$ are the hue channels of the reflectance map and the source image after converting them from the RGB to the HSV color space.

For the first term, as we know, low-light images always suffer from low contrast, which often indicates smaller gradient magnitudes~\cite{li2018structure}. Hence we attempt to manipulate the gradient magnitudes of the reflectance to boost the contrast of the enhanced results by amplifying the gradient of the input image with the factor $\beta$. Notice that $\nabla \hat{S}$ is a variant of the gradient of the input image $\nabla{S}$, in which small gradients, i.e. the noise, are suppressed before ampliﬁcation.
\begin{equation}
\nabla \hat{S}=
\begin{cases}
0& \text{if } |\nabla{S}|<\epsilon\\
\nabla{S}& \text{otherwise}
\end{cases}`
\end{equation}

The second term is adopted to avoid color mismatch in the decomposed reflectance, which enforce the hue component of the reflectance map to be close enough to that of the source image.

\textbf{Noise estimation loss.} Although the reflectance contrast loss attempts to suppress noise by ignoring small gradients, noise may still appear in flat dark regions. It is necessary to guide the decomposition of the noise map by,
\begin{equation}
    \mathcal{L}_{N} = \left\|S \times N\right\|_F
\end{equation}
which constrains the overall intensity of the noise~\cite{li2018structure}.

As for the detailed network configuration, SID-Net first adopts a commonly used feature extractor, which could be any classic network such as ResNet, U-Net, or even a simple CNN structure. Then, three separate $3\times3$ convolutional layers project the image features into branches, namely reflectance $R$, illumination $L$, and noise $N$ respectively. At the end, sigmoid function is used to constrain both $R$ and $L$ in the range of [0, 1], while tanh function is adopted to simulate additive noise to make $N$ fall in [-1, 1].

\subsection{NISM: Nonlinear Illumination Saturation Mapping Function}
\begin{figure}
\centering
\begin{minipage}[b]{0.3\linewidth}
  \centering
  \centerline{\includegraphics[width=1\linewidth]{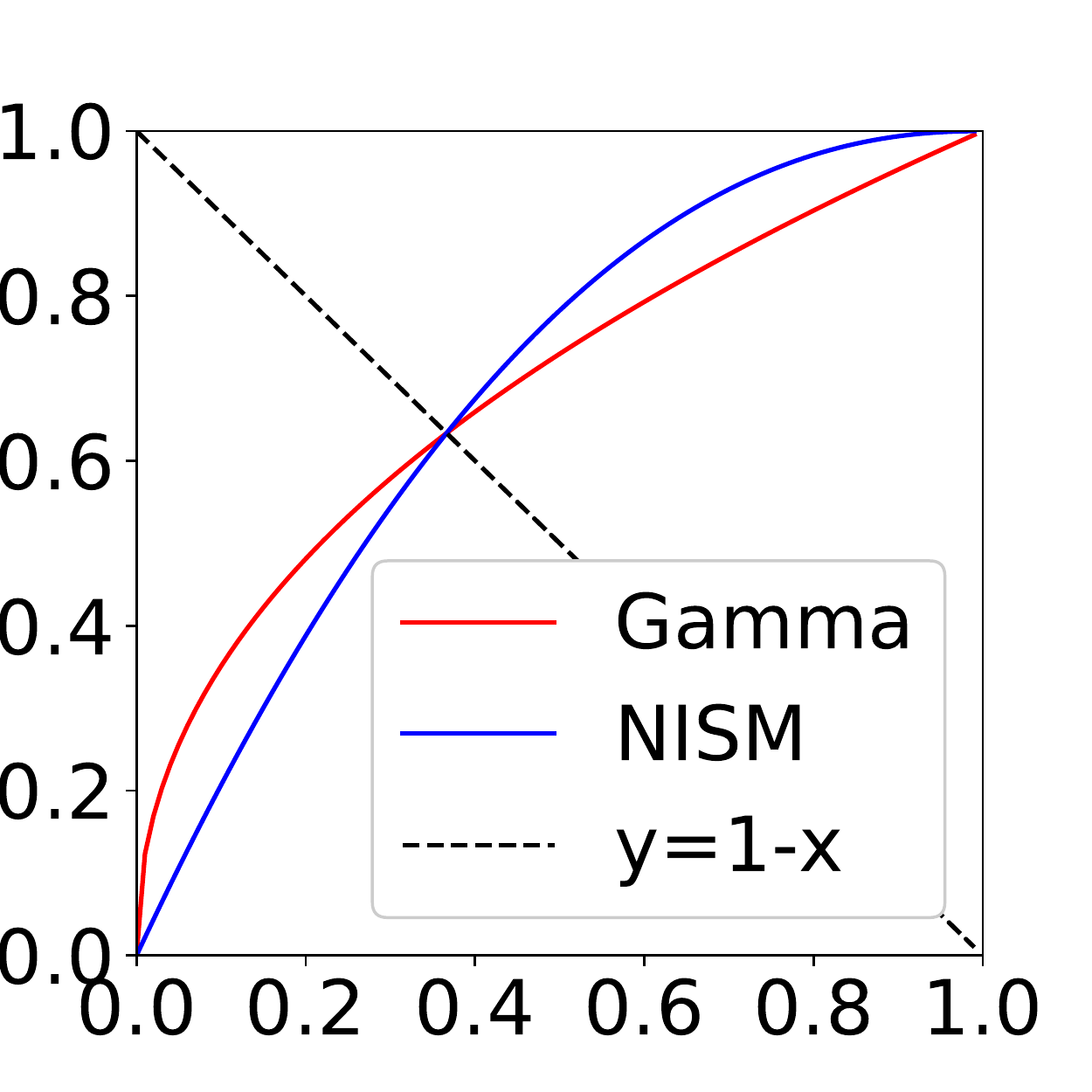}}
  \centerline{(a) Curves}\medskip
\end{minipage}
\begin{minipage}[b]{0.35\linewidth}
  \centering
  \centerline{\includegraphics[width=1\linewidth]{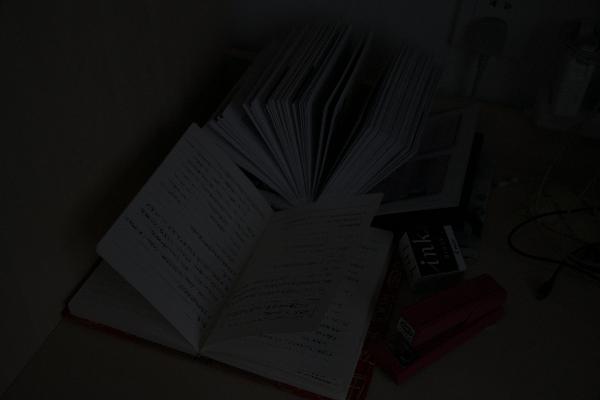}}
  \centerline{(b) Input}\medskip
\end{minipage}\\
\begin{minipage}[b]{0.35\linewidth}
  \centering
  \centerline{\includegraphics[width=1\linewidth]{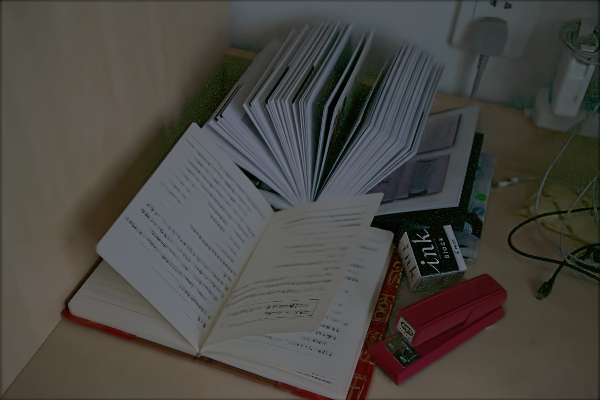}}
  \centerline{(c) Gamma}\medskip
\end{minipage}
\begin{minipage}[b]{0.35\linewidth}
  \centering
  \centerline{\includegraphics[width=1\linewidth]{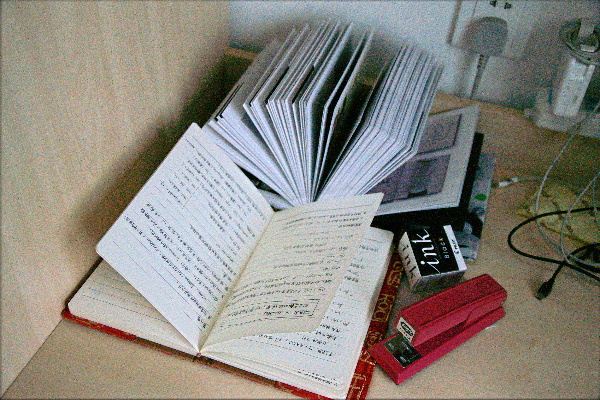}}
  \centerline{(d) NISM}\medskip
\end{minipage}
\caption{Comparison between the Gamma correction and the proposed manipulation operator NISM. (a) are the two function curves when $\gamma=2.2$ and $\eta=2.2$. (c) and (d) are the corresponding enhanced results when adopting Gamma correction and NISM on the illumination map of the low-light image (b) respectively.}
\label{fig:NISM}
\end{figure}

After obtaining the low-light illumination map $L$ of the given image $S$ by SID-Net, a manipulation operator would be adopted to enhance $L$ as described in Sect.~\ref{sect:retinex}. In the previous methods~\cite{guo2016lime,ren2018joint,zhang2018high}, researchers got used to refine the illumination map through Gamma correction, say $\hat{L}=L^{1/\gamma}$. When $\gamma>1$, the illumination map will be brightened up. In Fig.~\ref{fig:NISM}(a), the function curve of Gamma transformation with $\gamma=2.2$ is shown by the red line. It can be seen that pixels whose intensity are less than 20\% of the maximum illumination value, namely 0.2 in the scale of [0, 1], will be brightened to 50\%, while the pixels whose original intensity is large enough ($>80\%$) will keep the themselves basically unchanged. This property of Gamma correction would definitely do favor to enhance the illumination map especially in those dark regions, but it also brings two major flaws: 1) the contrast of the original image is destroyed due to the over-brightening of dark areas; 2) the illumination level of the final enhanced image is still insufficient because the bright areas are barely changed. As shown in Fig.~\ref{fig:NISM}(c), the overall quality of the enhanced result is improved compared with the low-light input Fig.~\ref{fig:NISM}(b), but it haven't been unable to reach the standard of 'normal-light' in human perception.

To combat the existing issues of Gamma correction, we consider depressing the brightening of dark pixels while increasing the illumination saturation of bright pixels. More concretely, for dark pixels, it is still necessary to be brightened up but should be carried on at a relatively lower level as compared with Gamma correction. In other words, the slope of the illumination operator at pixels whose intensity are less than 0.2 should be smaller than that of Gamma correction. On the contrary, the slope at bright pixels should be larger than that of Gamma correction, which could increase the illumination levels of bright pixels so as to render the final enhanced results bright enough.

According to the above considerations, a novel manipulation operator called Nonlinear Illumination Saturation Mapping (NISM) is proposed, which is formulated as,
\begin{equation}
    \hat{L}=1-(1-L)^\eta
\end{equation}
where $\eta$ is the control parameter as $\gamma$ in Gamma correction. The function curve of NISM with $\eta=2.2$ is depicted in the blue line in Fig.~\ref{fig:NISM}(a). It can be seen that the curve shape of NISM is consistent with the previous analysis about the improved curve slope. Indeed, the constructed NISM and the Gamma correction are symmetrical about $y=1-x$.

The final enhanced image is generated by recombining the decomposed reflectance $R$ and the refined illumination $\hat{L}$ by $R\times\hat{L}$. The corresponding result is shown in Fig.~\ref{fig:NISM}(d), which reflects that NISM could transform the decomposed illumination map of the original low-light image into a normal-light level.

In implementation $\eta$ is not fixed. It is computed by first clustering pixels of the low-light illumination map into two clusters, bright pixels and dark pixels, by KMeans. Then taking the minimum illumination value of bright pixels as threshold $T$, $\eta$ is calculated by,
\begin{equation}
    \eta=\log(1-0.8)/\log(1-T)
\end{equation}
which means to map the minimum illumination value of bright pixels to 0.8 under NISM.

\section{Experimental Results}
\subsection{Experiment Settings}
As a self-supervised framework, SID-NISM doesn't need any training dataset or prior information. For any given low-light image, SID-Net could split it into reflectance, illumination and noise components directly within hundred iterations, then NISM would enhance the decomposed illumination to combine with the reflectance. Besides, all the coefficients mentioned in Sect. \ref{sect:sid-nism} would remain the same during the experiments, which means it is not necessary to adjust the parameters for different inputs.

Evaluations are performed on real-scene images from four public datasets, DICM~\cite{lee2013contrast}, LIME~\cite{guo2016lime}, LOL~\cite{wei2018deep}, and MEF~\cite{ma2015perceptual}. And the proposed method SID-NISM is compared with 6 state-of-the-art low-light image enhancement algorithms, including 1) a weighted variational model for simultaneous reflectance and illumination estimation (SRIE)~\cite{fu2016weighted}, 2)illumination map estimation method (LIME)~\cite{guo2016lime}, 3) joint enhancement and denoising method via sequential decomposition (JED)~\cite{ren2018joint}, 4) perceptually bidirectional similarity based illumination estimation (PBS)~\cite{zhang2018high}, 5) deep learning based Retinex decomposition (Retinex-Net)~\cite{wei2018deep}, and 6) unsupervised-learning based back-lit image restoration network (ExCNet)~\cite{zhang2019zero}.

In the following subsections, both subjective and objective experimental results will be exhibited. More experiments including the intermediate decomposition results and the analysis of the effect on the object detection task can be found in the supplementary materials.

\subsection{Image Decomposition Results}
\begin{figure*}
\centering
\begin{minipage}[b]{0.2\linewidth}
  \centering
  \centerline{\includegraphics[width=1\linewidth]{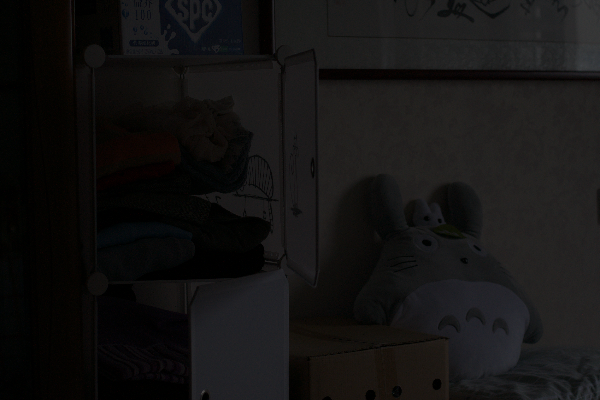}}
\end{minipage}
\begin{minipage}[b]{0.2\linewidth}
  \centering
  \centerline{\includegraphics[width=1\linewidth]{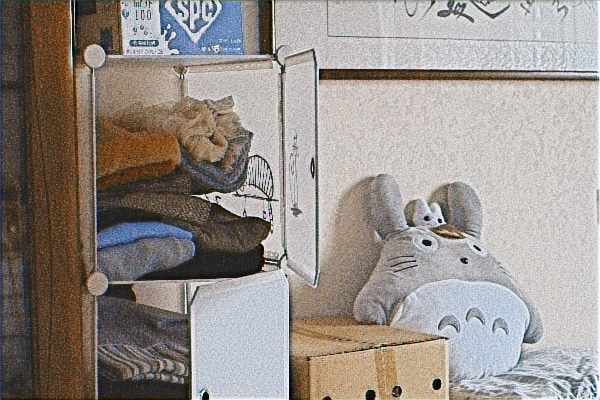}}
\end{minipage}
\begin{minipage}[b]{0.2\linewidth}
  \centering
  \centerline{\includegraphics[width=1\linewidth]{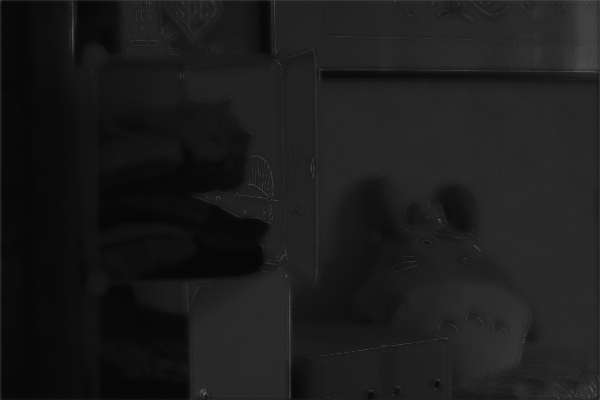}}
\end{minipage}
\begin{minipage}[b]{0.2\linewidth}
  \centering
  \centerline{\includegraphics[width=1\linewidth]{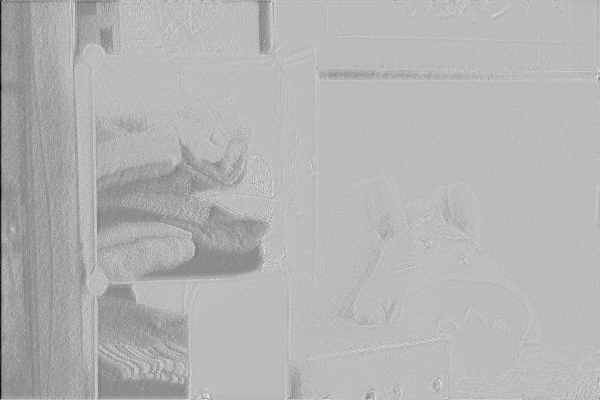}}
\end{minipage}\\
\begin{minipage}[b]{0.2\linewidth}
  \centering
  \centerline{\includegraphics[width=1\linewidth]{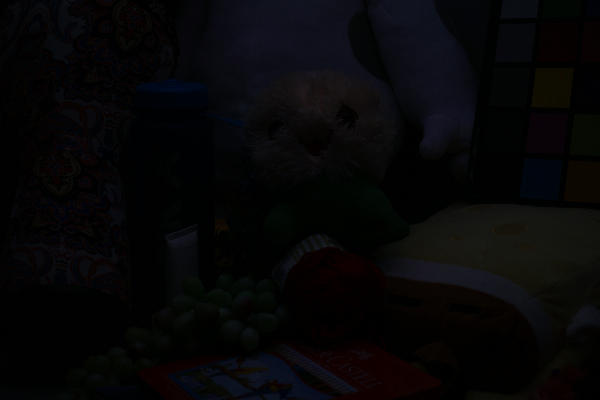}}
  \centerline{(a) Source $S$}\medskip
\end{minipage}
\begin{minipage}[b]{0.2\linewidth}
  \centering
  \centerline{\includegraphics[width=1\linewidth]{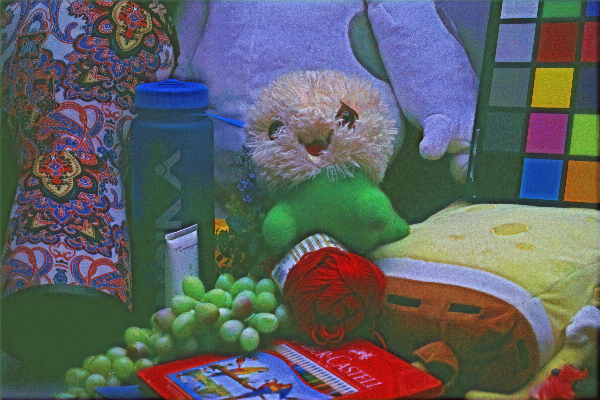}}
  \centerline{(b) Reflectance $R$}\medskip
\end{minipage}
\begin{minipage}[b]{0.2\linewidth}
  \centering
  \centerline{\includegraphics[width=1\linewidth]{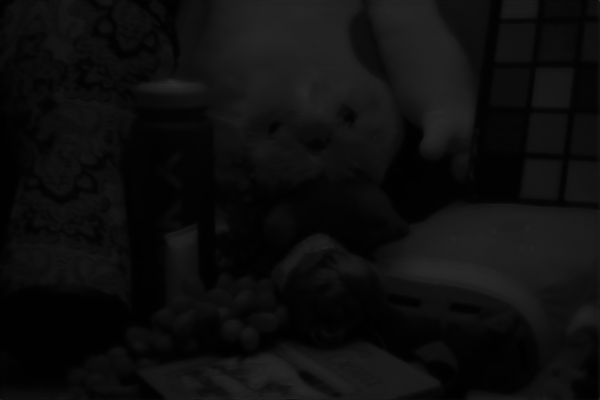}}
  \centerline{(c) Illumination $L$}\medskip
\end{minipage}
\begin{minipage}[b]{0.2\linewidth}
  \centering
  \centerline{\includegraphics[width=1\linewidth]{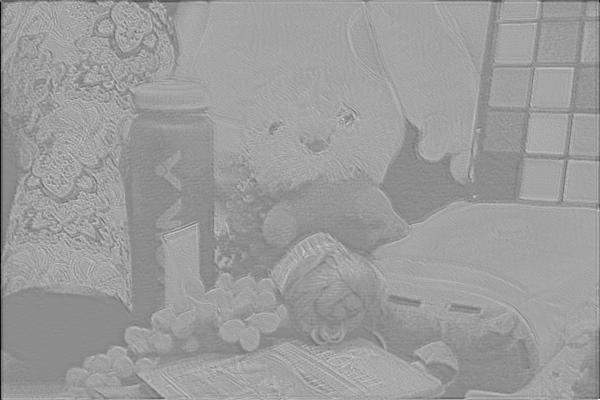}}
  \centerline{(d) Noise $N$}\medskip
\end{minipage}
\caption{Intermediate decomposition results of two low-light images, \textit{Totoro} and \textit{toy} in LOL dataset.}
\label{fig:decomposition1}
\end{figure*}

Since Retinex-based low-light image enhancement methods consider the problem as an image decomposition problem, it is necessary to show our intermediate decomposition results firstly. Fig.~\ref{fig:decomposition1} exhibits the decomposition results of two low-light images, \textit{Totoro} and \textit{toy} in LOL dataset. Their final enhanced results can be found in the main paper. In general, the proposed SID-Net can decompose reasonable reflectance, illumination and noise maps for given input images in a self-supervised way successfully. The decomposed reflectance maps contain sufficient color and structure details of the captured scene. The illumination maps reflect the lighting conditions as observed in human eyes. While the noise maps extract underlying noise component in dark regions.

\begin{figure*}[t]
\centering
\includegraphics[scale=0.4]{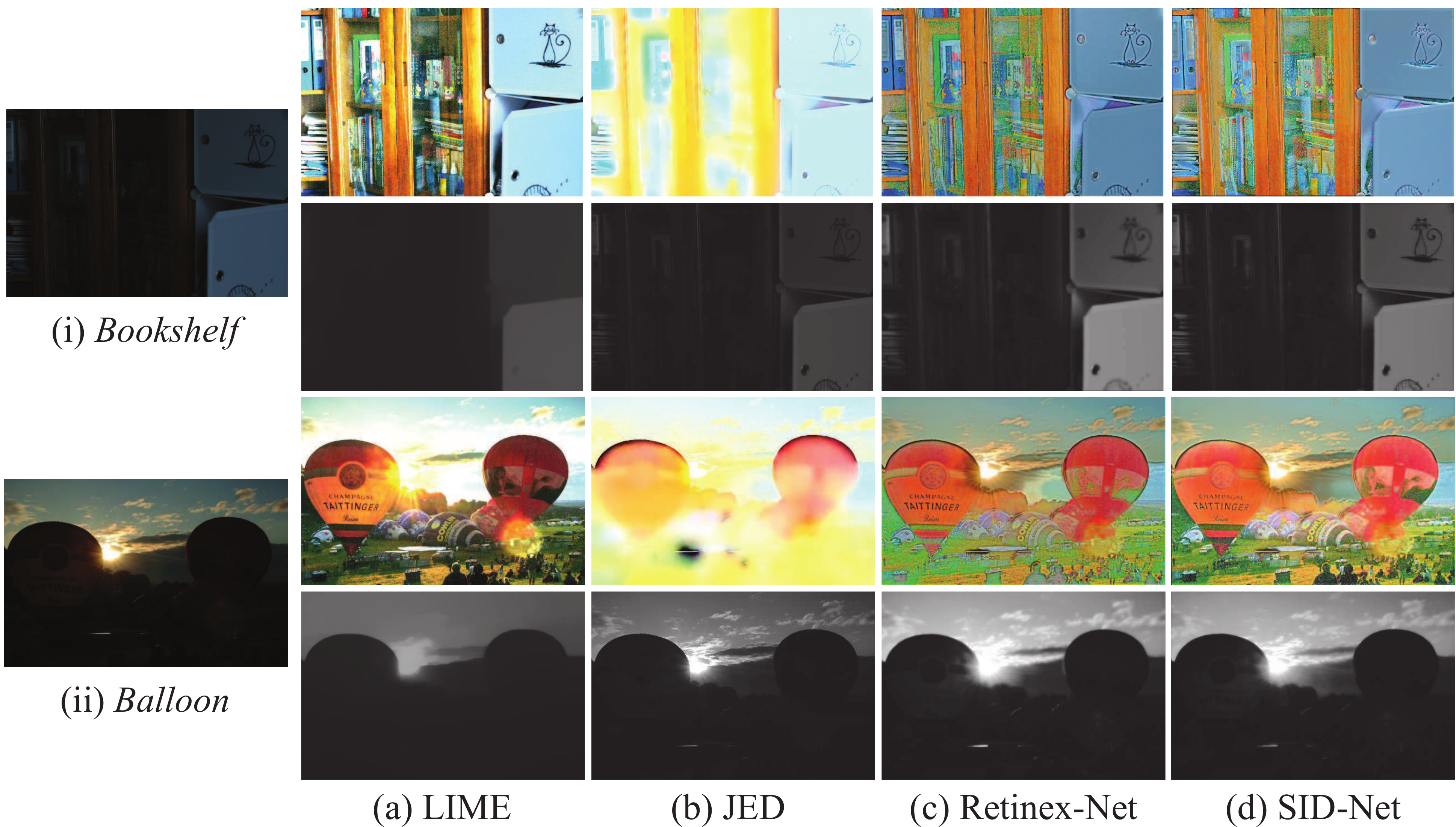}
\caption{The decomposition results of LIME, JED, Retinex-Net, and SID-Net for \textit{Bookshelf} in LOL dataset and \textit{Balloon} in MEF dataset. For each of the example, the first row are the decomposed reflectance, while the second are the illumination maps}
\label{fig:decomposition2}
\end{figure*}

Fig.~\ref{fig:decomposition2} compares the decomposition results of the proposed image decomposition network SID-Net with three state-of-the-art Retinex-based methods, LIME, JED and Retinex-Net, on \textit{Bookshelf} in LOL dataset and \textit{Balloon} in MEF dataset.

It can be seen that all of the four methods can decompose reasonable reflectance and illumination maps for given input images as expected. However, there also exists some differences among the four methods. In the aspect of reflectance, it is obvious that LIME can restore more color and structure details of the original objects compared with the other methods, because the reflectance of LIME is calculated by conducting element-wise division on the source image and the estimated illumination, which preserve the Retinex theory to a large extent. As for the illumination maps, SID-Net and Retinex-Net outperform other methods due to the specific structure-awareness illumination smoothness loss term, which makes it more consistent with the image structure. Besides, both as image decomposition networks, the biggest difference between SID-Net and Retinex-Net is that the proposed SID-Net is an unsupervised learning network without any prior training process, while the other one is a supervised learning network based on large-scale training dataset. As for the practical image decomposition performance, the two networks are evenly matched. Even in the reflectance maps, the results of SID-Net are clearer with more details than that of Retinex-Net.
In general, as shown in Fig.~\ref{fig:decomposition2}, SID-Net can extract underlying consistent reflectance from the given input image directly, and portray the lightness and shadow of the image in the illumination map. Although without prior training, the SID-Net can achieve the same or even better performance at the image decomposition stage.

\subsection{Human Perception User Study}\label{sect:userstudy}
\begin{figure}
\centering
\begin{minipage}[b]{0.29\linewidth}
  \centering
  \centerline{\includegraphics[width=1\linewidth]{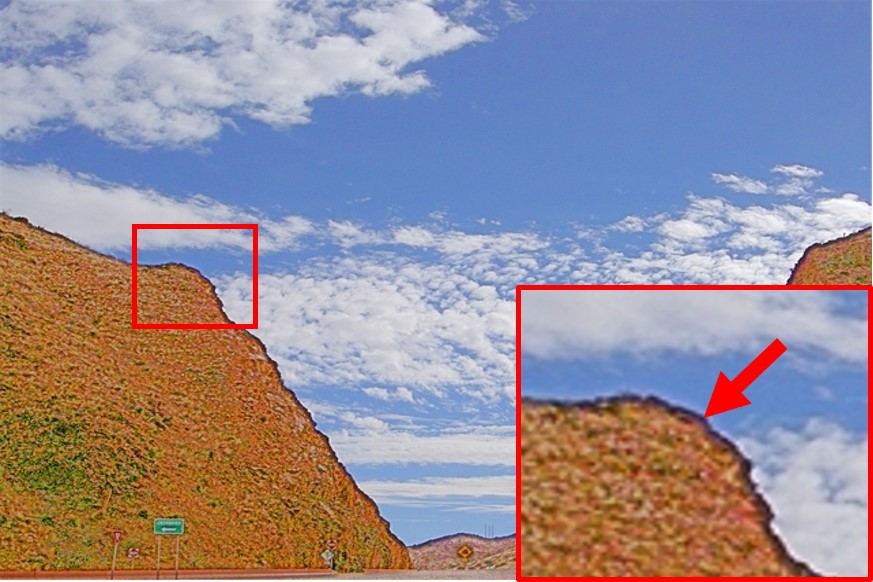}}
  \centerline{(a) Black Edges}\medskip
\end{minipage}
\begin{minipage}[b]{0.29\linewidth}
  \centering
  \centerline{\includegraphics[width=1\linewidth]{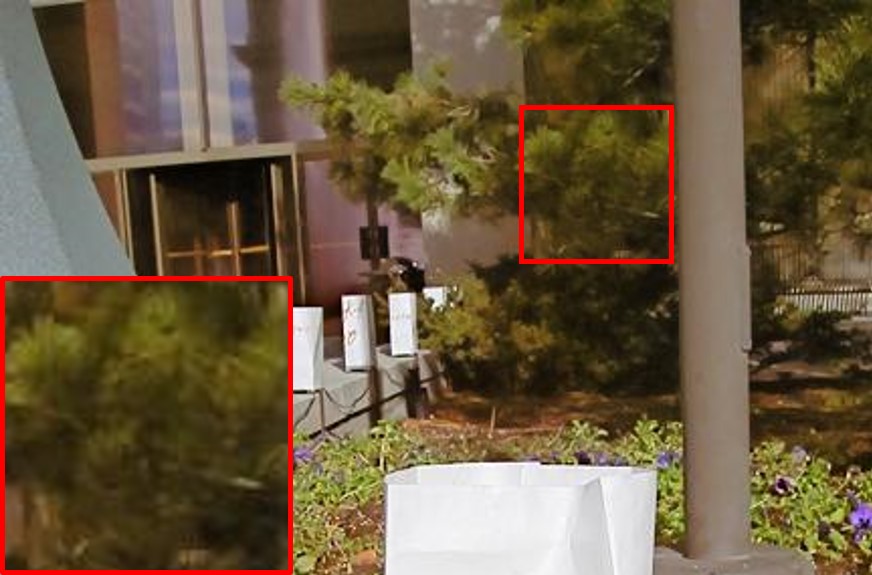}}
  \centerline{(b) Blur}\medskip
\end{minipage}\\
\begin{minipage}[b]{0.29\linewidth}
  \centering
  \centerline{\includegraphics[width=1\linewidth]{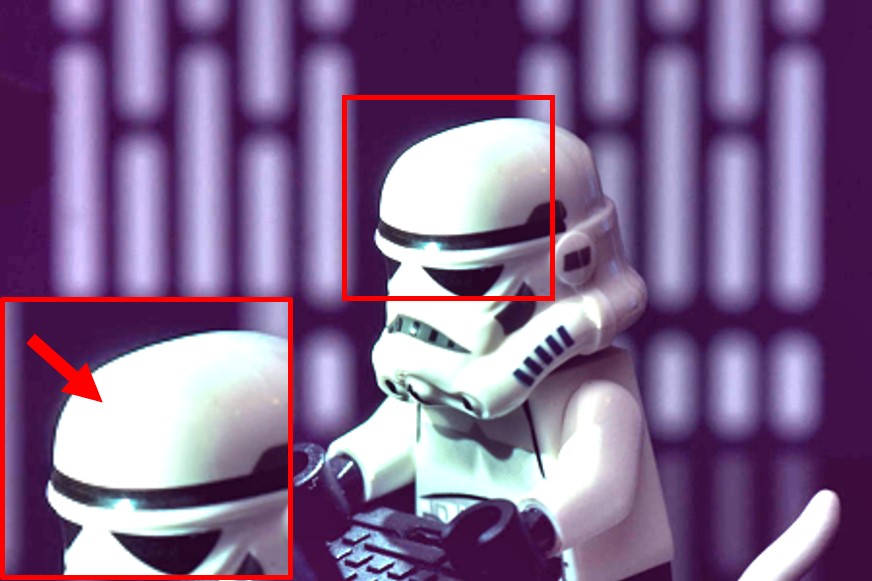}}
  \centerline{(c) Overexposure}\medskip
\end{minipage}
\begin{minipage}[b]{0.29\linewidth}
  \centering
  \centerline{\includegraphics[width=1\linewidth]{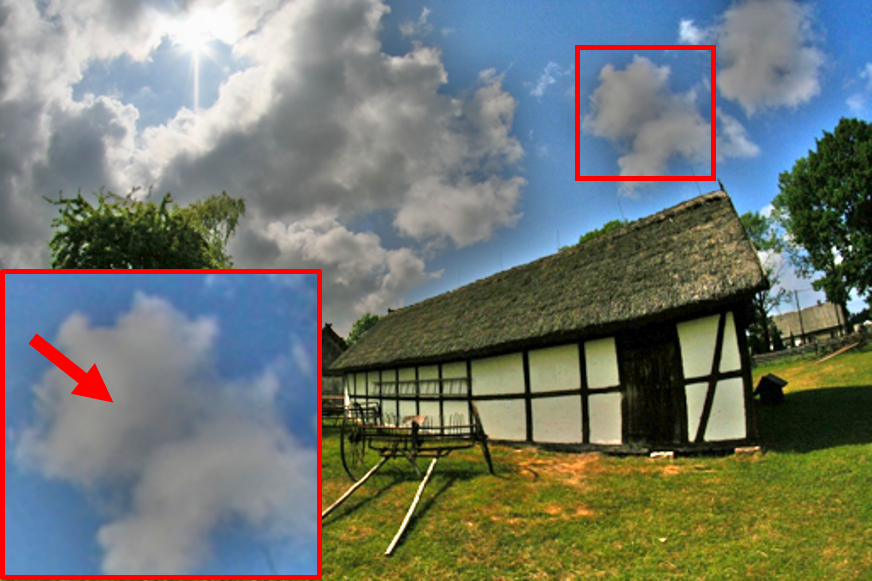}}
  \centerline{(d) Gray Shadow}\medskip
\end{minipage}
\begin{minipage}[b]{0.29\linewidth}
  \centering
  \centerline{\includegraphics[width=1\linewidth]{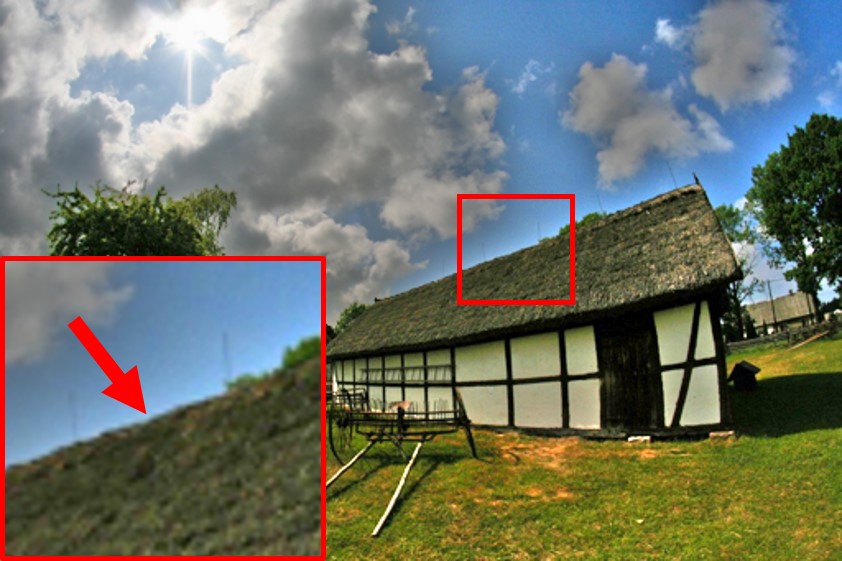}}
  \centerline{(e) Halo}\medskip
\end{minipage}
\caption{Five artifacts examples from (a) Retinex-Net \cite{wei2018deep} on \textit{hill} in DICM dataset; (b) JED \cite{ren2018joint} on \textit{tree} in DICM dataset; (c) LIME \cite{guo2016lime} on \textit{stormtrooper} in LIME dataset; (d) and (e) ExCNet \cite{zhang2019zero} on \textit{roof} in MEF dataset.}
\label{fig:artifacts}
\end{figure}

User study was conducted to perform comparisons between state-of-the-art methods in the dimensions of the overall image quality and some unpleasing artifacts through human perceptions. To evaluate the overall image quality, the subjects gave their personal feelings on the scale of 0 to 10, where 0 represents ``poor'', 5 means ``good'', and 10 is ``excellent''. As for the artifacts, there are five artifacts to be selected from (multiple-choice). Those are black edges, blur, overexposure, gray shadow and halo as shown in Fig.~\ref{fig:artifacts}. Once the subjects observe phenomenon in the images consisting with the description of the artifacts, the corresponding option would be checked. For all the enhanced images obtained by different methods, the subjects move the slider to give the image quality score and select specific artifacts according to their observations. The above procedure was completed on the Amazon Mechanical Turk platform by thirty workers.

\begin{figure}
\centering
\includegraphics[scale=0.32]{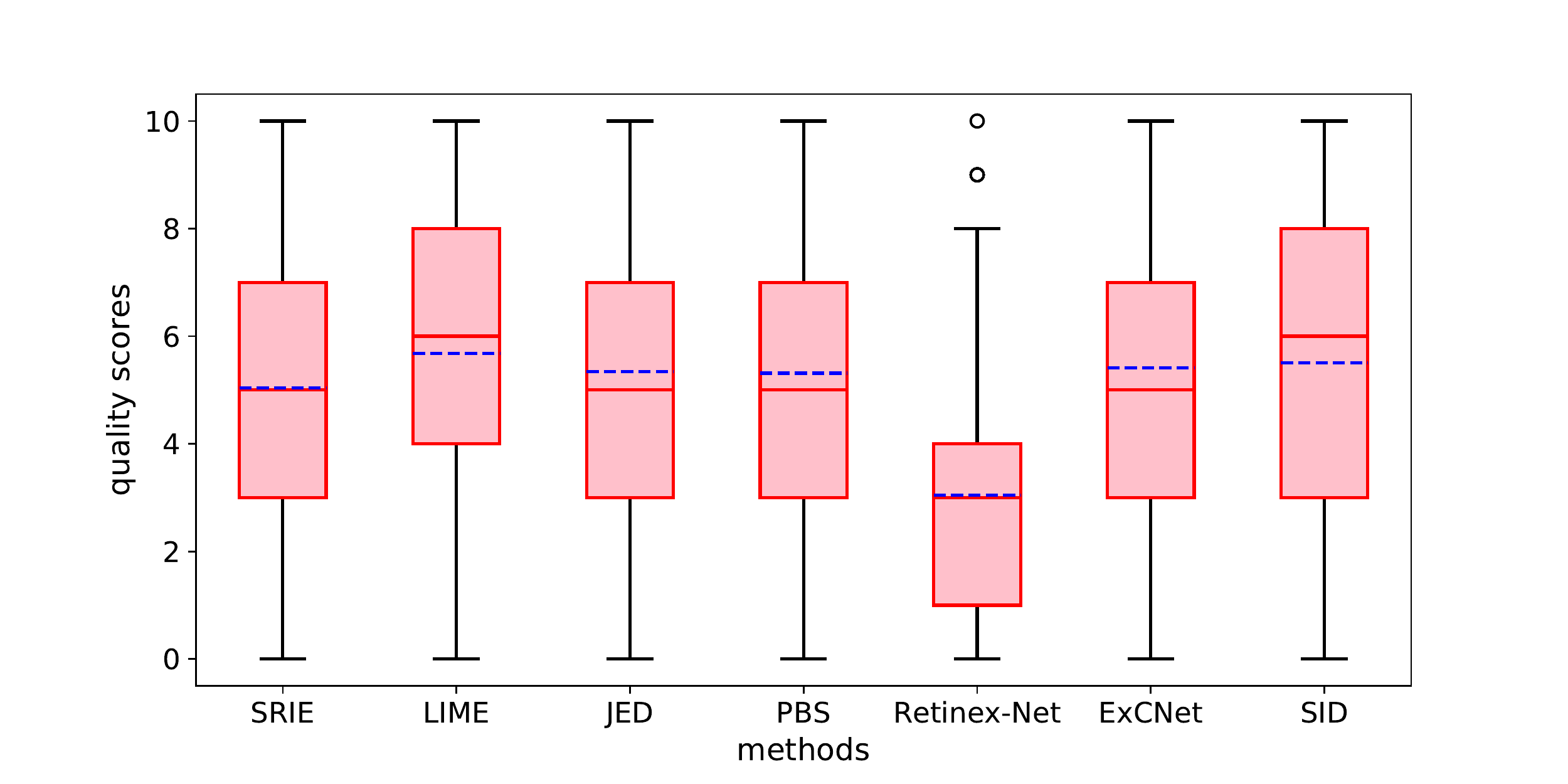}
\caption{The image quality scores for different methods.}
\label{fig:qualityUser}
\end{figure}

\begin{figure}
\centering
\includegraphics[scale=0.32]{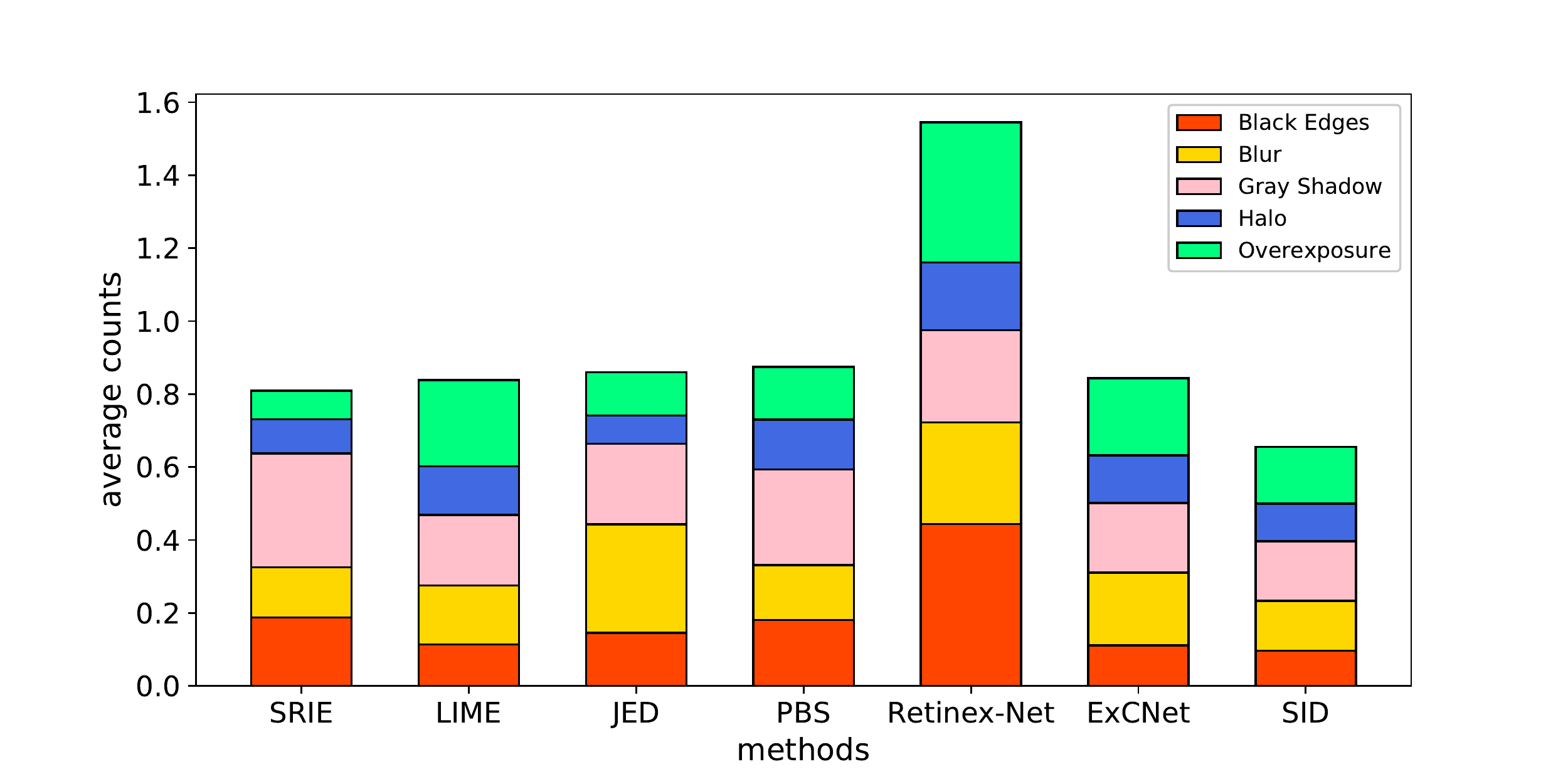}
\caption{The average counts of artifacts in different methods.}
\label{fig:artifactsUser}
\end{figure}

The results of the user study are summarized in graphs shown in Fig.~\ref{fig:qualityUser} and \ref{fig:artifactsUser}. In Fig.~\ref{fig:qualityUser}, a boxplot is used to exhibit the distributions of the image quality scores for different methods, in which the red solid line represents the median while the blue dotted line represents the mean score. Through observing the median, mean, and the first quartile, it is obvious that the participants showed a strong bias in preference towards our results (median: 6.0, mean: 5.51, quartile: 8.0)  when compared to Retinex-Net~\cite{wei2018deep} (median: 3.0, mean: 3.05, quartile: 4.0), gave higher scores when compared to SRIE~\cite{fu2016weighted} (median: 5.0, mean: 5.04, quartile: 7.0), JED~\cite{ren2018joint} (median: 5.0, mean: 5.34, quartile: 7.0), PBS~\cite{zhang2018high} (median: 5.0, mean: 5.31, quartile: 7.0) and ExCNet~\cite{zhang2019zero} (median: 5.0, mean: 5.41, quartile: 7.0), and had no preference when compared with LIME~\cite{guo2016lime} (median: 6.0, mean: 5.67, quartile: 8.0). The results in Fig.~\ref{fig:qualityUser} indicate that SID-NISM outperforms most of the state-of-the-art methods in terms of the overall image quality.

In Fig.~\ref{fig:artifactsUser}, a stacked bar graph is adopted to illustrate the average counts of artifacts in the results of different methods, in which each color bar corresponds to the five artifacts in Fig.~\ref{fig:artifacts} respectively. Specifically, the average counts are computed by $M/(N*P)$, where $M$ is the total counts of artifacts in the results of each method, $N$ is the number of images, $P$ is the number of subjects. It can be seen that SID-NISM tends to introduce less artifacts (average counts: 0.6) than other methods (average counts: above 0.8) in total. Generally, SID-NISM doesn't lead too many unexpected artifacts. SRIE~\cite{fu2016weighted} would introduce gray shadow into their results, while the results of LIME~\cite{guo2016lime} may be over-enhanced to become overexposure. JED~\cite{ren2018joint} has more blurry cases, while Retinex-Net~\cite{wei2018deep} contains the most black edges in their results. These phenomenons can be also observed in the visual examples of the next subsection.

\subsection{Visual Quality Comparisons}\label{sect:visualQuality}
\begin{figure*}
\centering
\begin{minipage}[b]{0.16\linewidth}
  \centering
  \centerline{\includegraphics[width=1\linewidth]{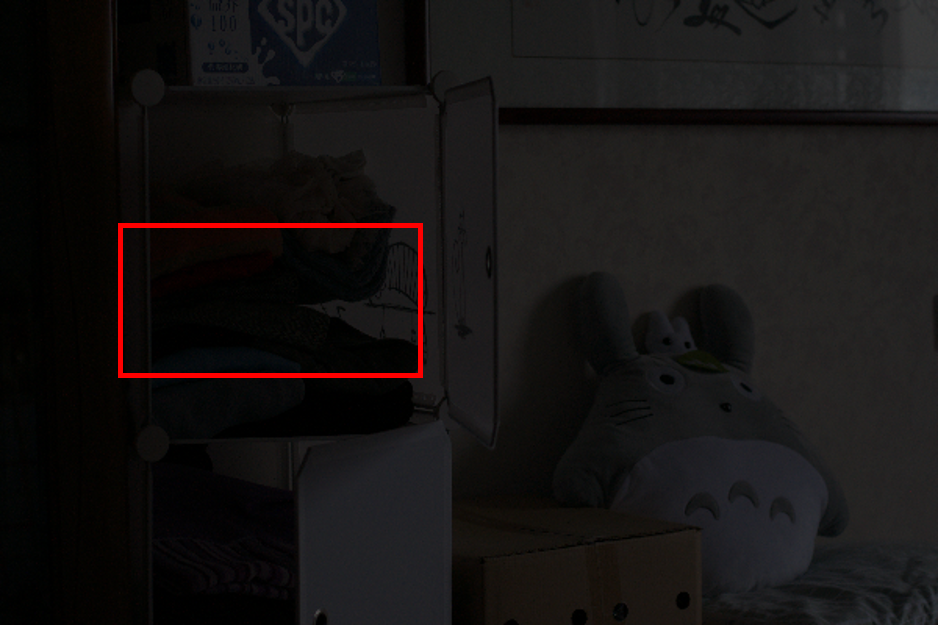}}
\end{minipage}
\begin{minipage}[b]{0.16\linewidth}
  \centering
  \centerline{\includegraphics[width=1\linewidth]{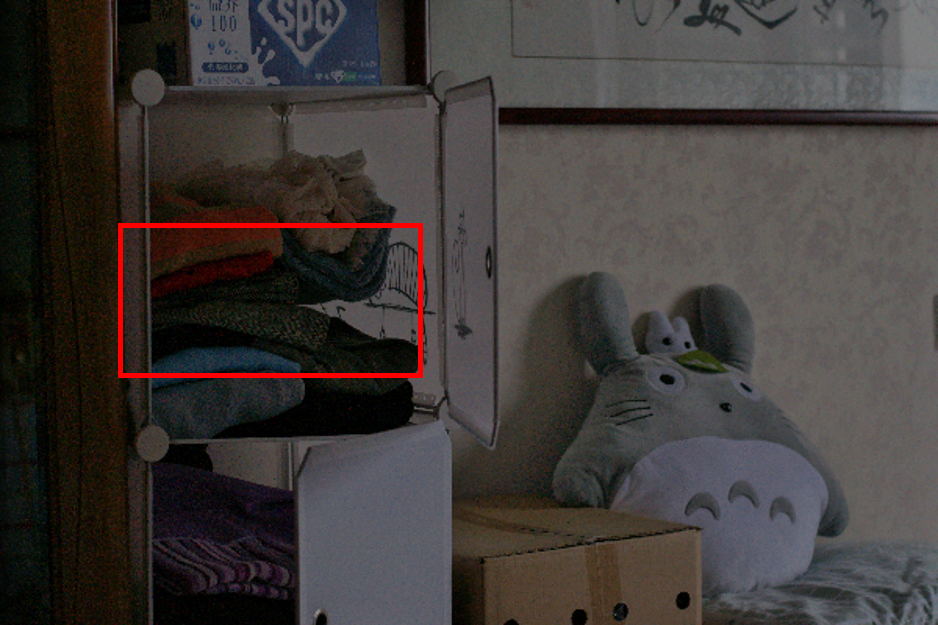}}
\end{minipage}
\begin{minipage}[b]{0.16\linewidth}
  \centering
  \centerline{\includegraphics[width=1\linewidth]{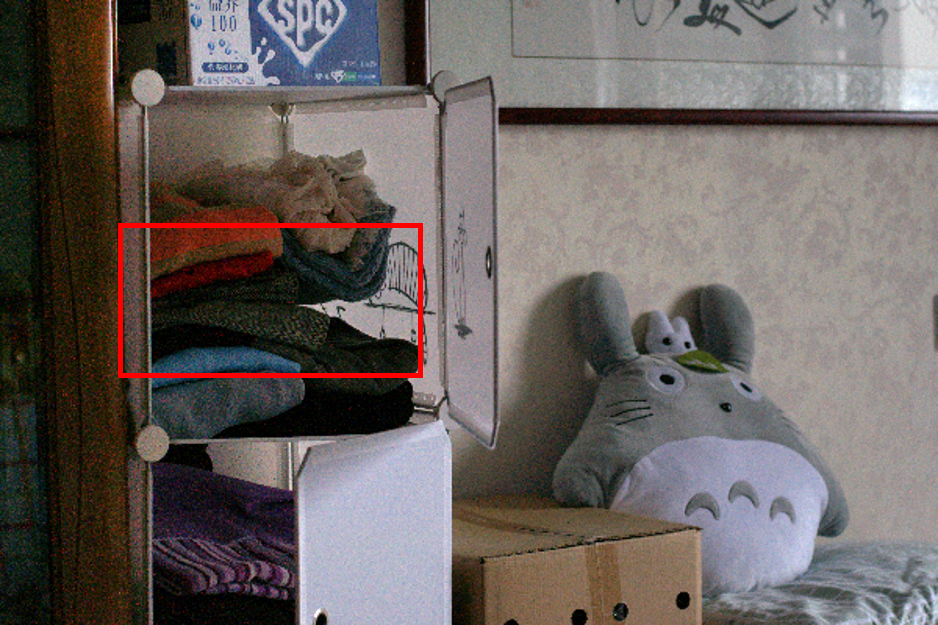}}
\end{minipage}
\begin{minipage}[b]{0.16\linewidth}
  \centering
  \centerline{\includegraphics[width=1\linewidth]{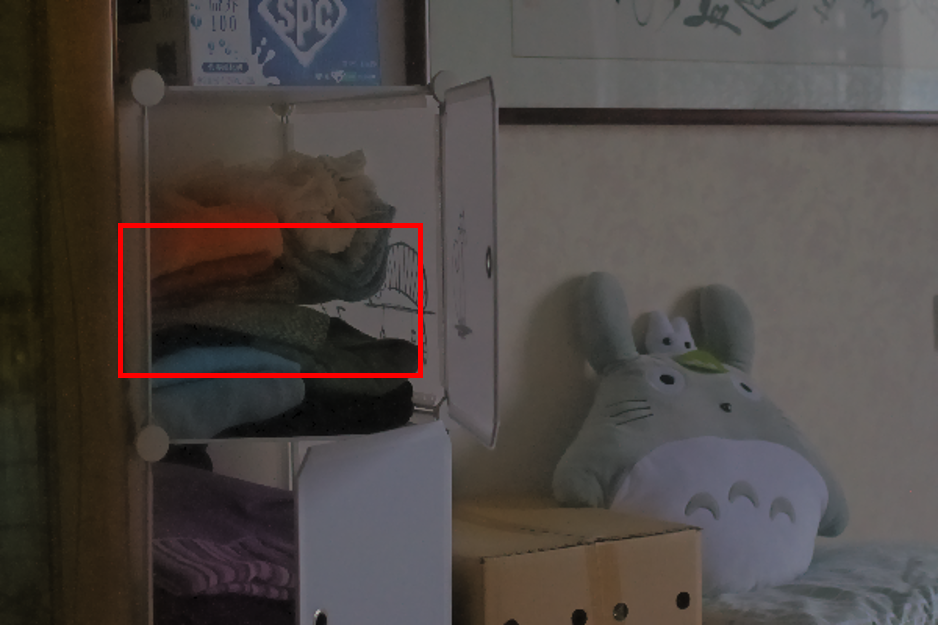}}
\end{minipage}
\begin{minipage}[b]{0.16\linewidth}
  \centering
  \centerline{\includegraphics[width=1\linewidth]{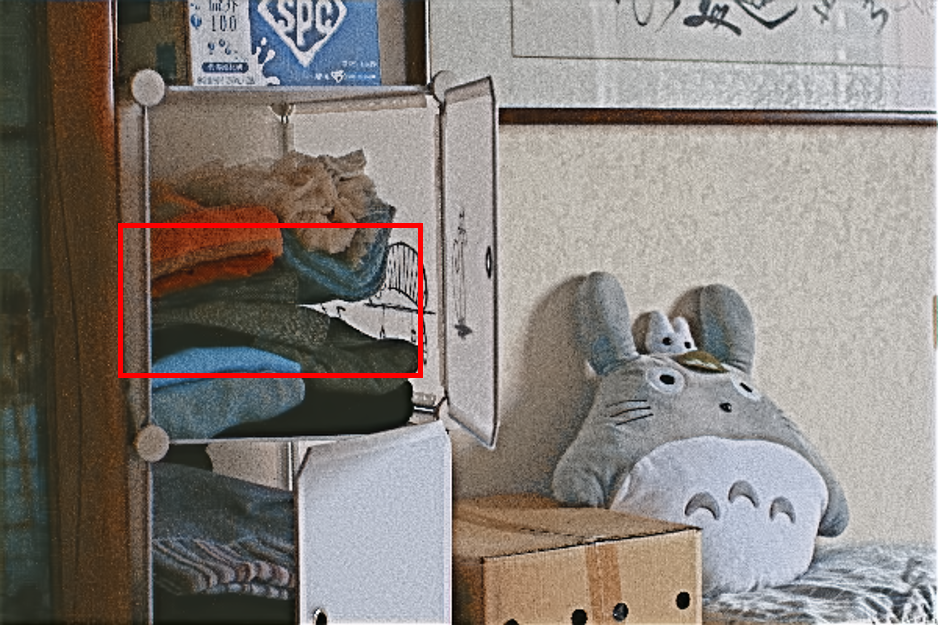}}
\end{minipage}\\
\begin{minipage}[b]{0.16\linewidth}
  \centering
  \centerline{\includegraphics[width=1\linewidth]{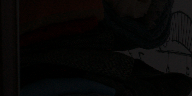}}
\end{minipage}
\begin{minipage}[b]{0.16\linewidth}
  \centering
  \centerline{\includegraphics[width=1\linewidth]{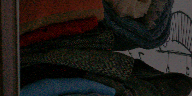}}
\end{minipage}
\begin{minipage}[b]{0.16\linewidth}
  \centering
  \centerline{\includegraphics[width=1\linewidth]{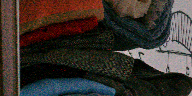}}
\end{minipage}
\begin{minipage}[b]{0.16\linewidth}
  \centering
  \centerline{\includegraphics[width=1\linewidth]{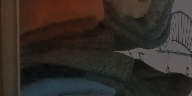}}
\end{minipage}
\begin{minipage}[b]{0.16\linewidth}
  \centering
  \centerline{\includegraphics[width=1\linewidth]{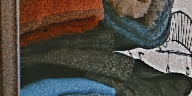}}
\end{minipage}\\
\begin{minipage}[b]{0.16\linewidth}
  \centering
  \centerline{\includegraphics[width=1\linewidth]{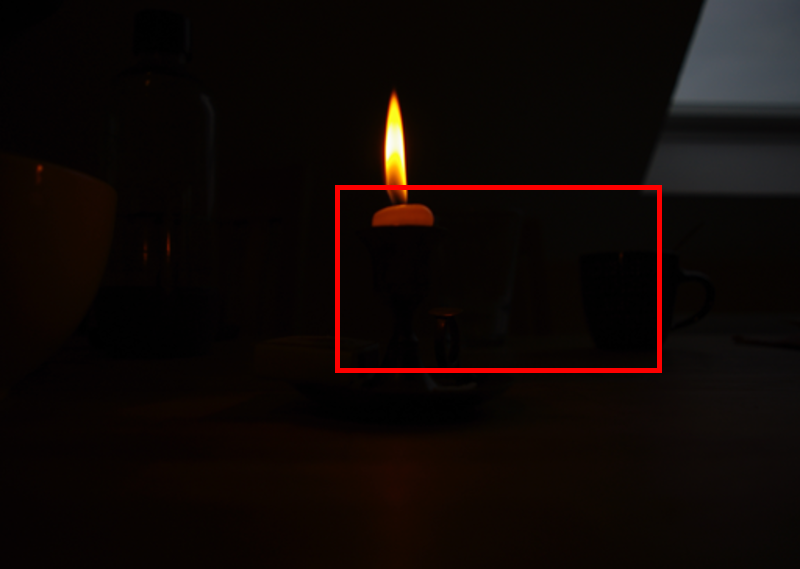}}
\end{minipage}
\begin{minipage}[b]{0.16\linewidth}
  \centering
  \centerline{\includegraphics[width=1\linewidth]{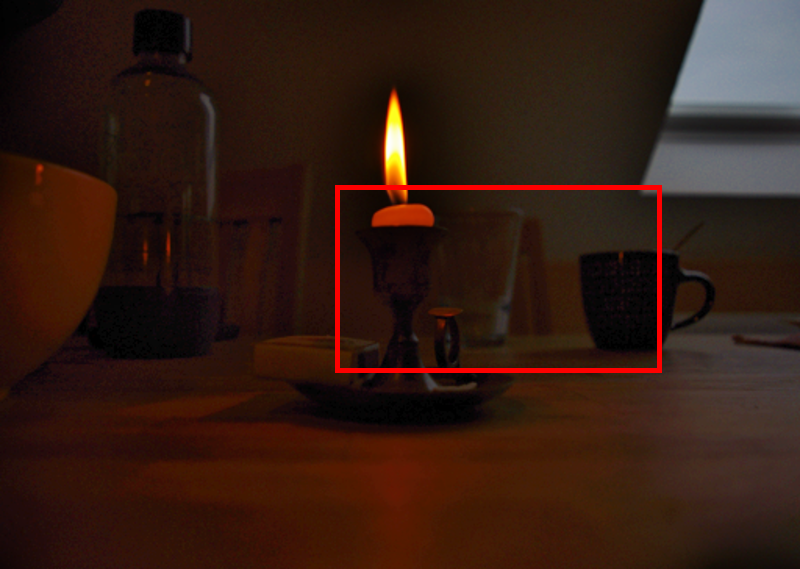}}
\end{minipage}
\begin{minipage}[b]{0.16\linewidth}
  \centering
  \centerline{\includegraphics[width=1\linewidth]{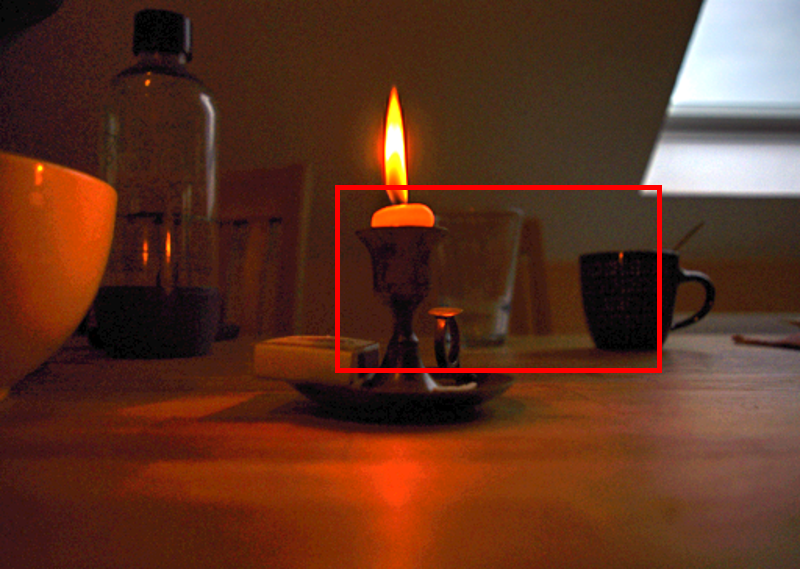}}
\end{minipage}
\begin{minipage}[b]{0.16\linewidth}
  \centering
  \centerline{\includegraphics[width=1\linewidth]{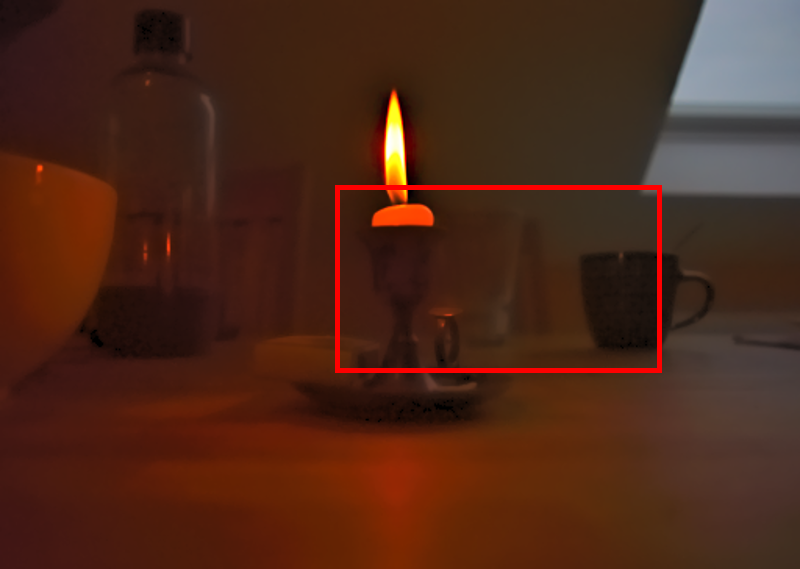}}
\end{minipage}
\begin{minipage}[b]{0.16\linewidth}
  \centering
  \centerline{\includegraphics[width=1\linewidth]{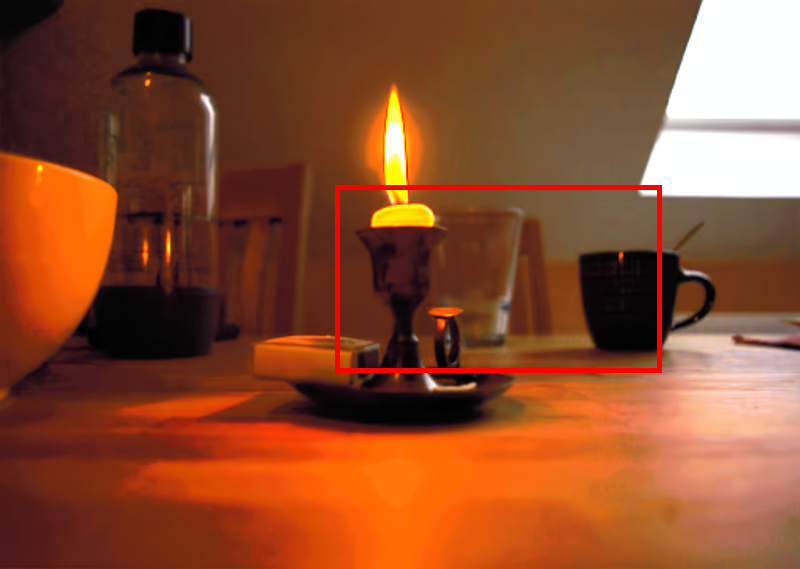}}
\end{minipage}\\
\begin{minipage}[b]{0.16\linewidth}
  \centering
  \centerline{\includegraphics[width=1\linewidth]{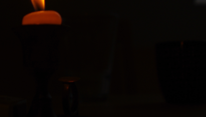}}
  \centerline{(a) Input}\medskip
\end{minipage}
\begin{minipage}[b]{0.16\linewidth}
  \centering
  \centerline{\includegraphics[width=1\linewidth]{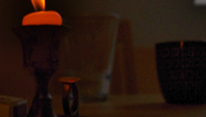}}
  \centerline{(b) SRIE}\medskip
\end{minipage}
\begin{minipage}[b]{0.16\linewidth}
  \centering
  \centerline{\includegraphics[width=1\linewidth]{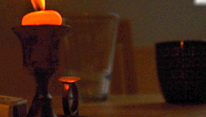}}
  \centerline{(c) LIME}\medskip
\end{minipage}
\begin{minipage}[b]{0.16\linewidth}
  \centering
  \centerline{\includegraphics[width=1\linewidth]{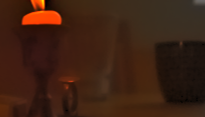}}
  \centerline{(d) JED}\medskip
\end{minipage}
\begin{minipage}[b]{0.16\linewidth}
  \centering
  \centerline{\includegraphics[width=1\linewidth]{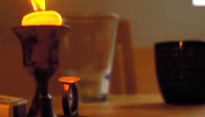}}
  \centerline{(e) SID-NISM}\medskip
\end{minipage}
\caption{Compared with the results obtained by SRIE \cite{fu2016weighted}, LIME \cite{guo2016lime} and JED \cite{ren2018joint} on two natural images: \textit{Totoro} from LOL dataset and \textit{candle} from MEF dataset. Below are the enlarged detailed images in the red box. (Best viewed on screen)}
\label{fig:vision1}
\end{figure*}

\begin{figure*}
\centering
\begin{minipage}[b]{0.16\linewidth}
  \centering
  \centerline{\includegraphics[width=1\linewidth]{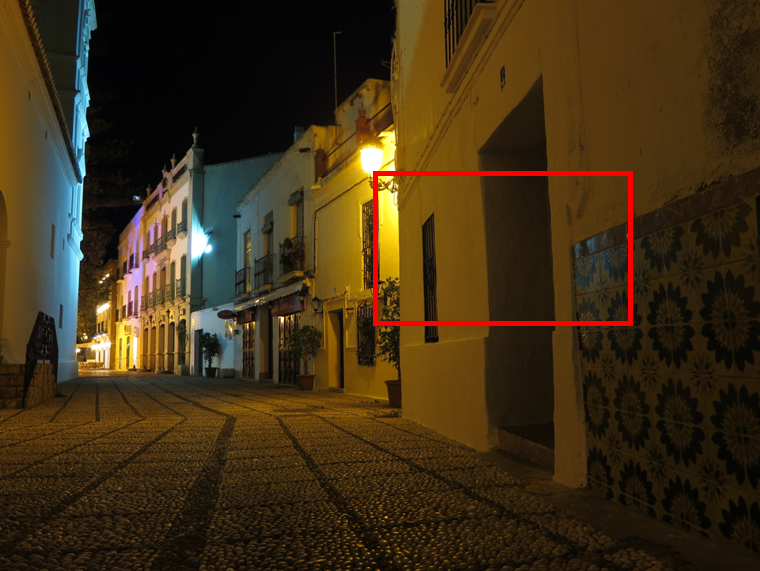}}
\end{minipage}
\begin{minipage}[b]{0.16\linewidth}
  \centering
  \centerline{\includegraphics[width=1\linewidth]{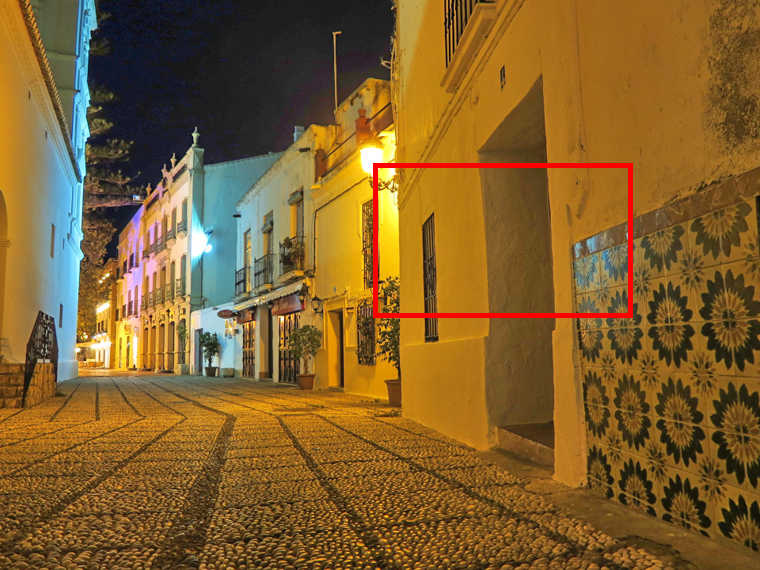}}
\end{minipage}
\begin{minipage}[b]{0.16\linewidth}
  \centering
  \centerline{\includegraphics[width=1\linewidth]{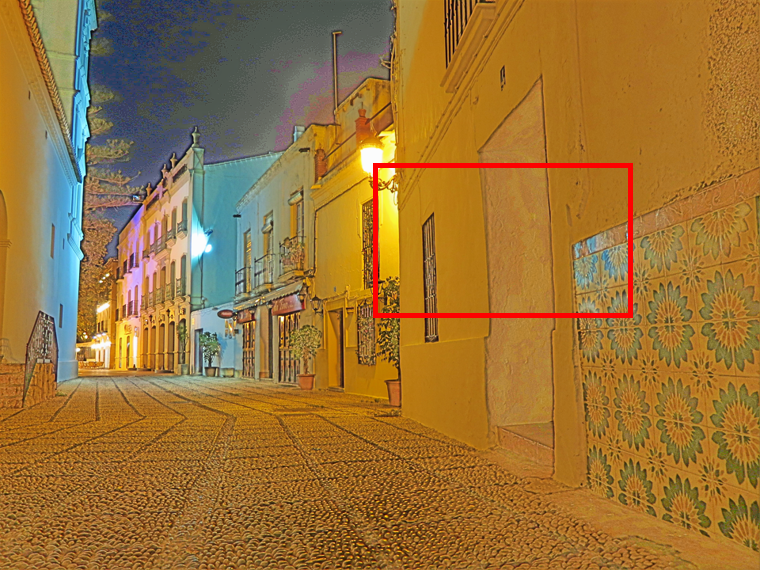}}
\end{minipage}
\begin{minipage}[b]{0.16\linewidth}
  \centering
  \centerline{\includegraphics[width=1\linewidth]{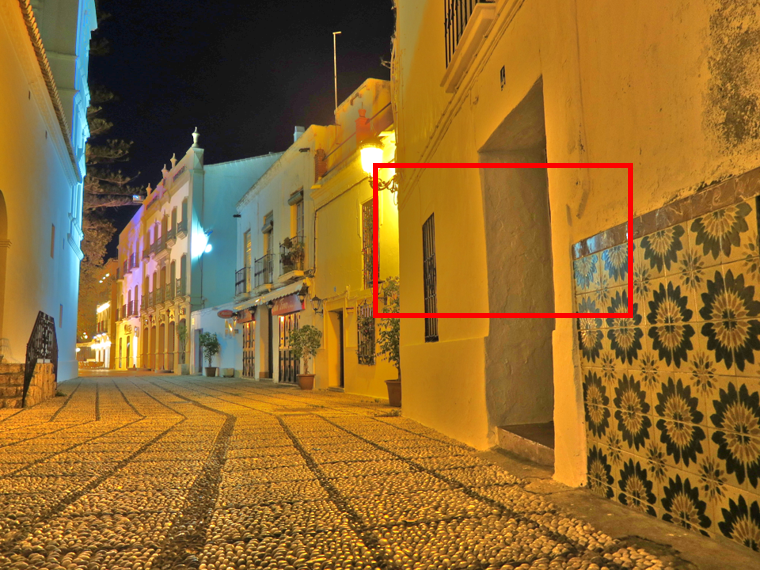}}
\end{minipage}
\begin{minipage}[b]{0.16\linewidth}
  \centering
  \centerline{\includegraphics[width=1\linewidth]{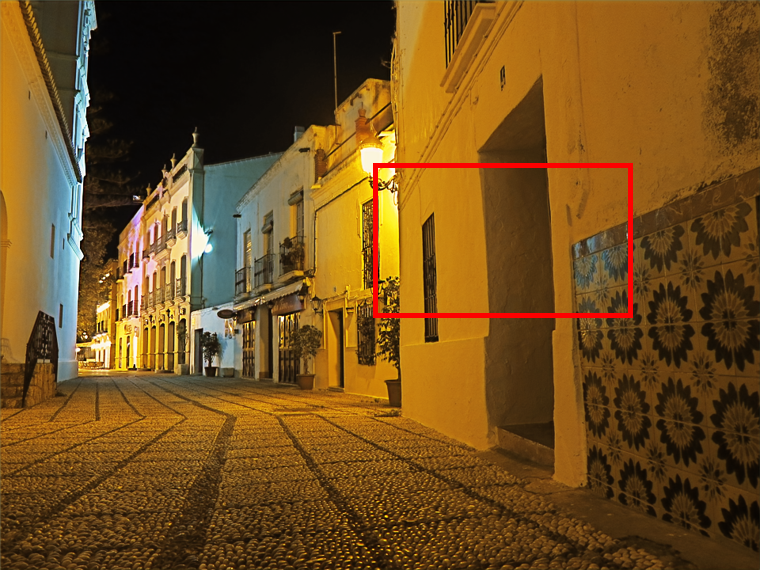}}
\end{minipage}\\
\begin{minipage}[b]{0.16\linewidth}
  \centering
  \centerline{\includegraphics[width=1\linewidth]{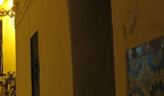}}
\end{minipage}
\begin{minipage}[b]{0.16\linewidth}
  \centering
  \centerline{\includegraphics[width=1\linewidth]{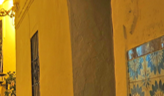}}
\end{minipage}
\begin{minipage}[b]{0.16\linewidth}
  \centering
  \centerline{\includegraphics[width=1\linewidth]{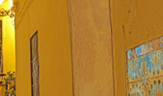}}
\end{minipage}
\begin{minipage}[b]{0.16\linewidth}
  \centering
  \centerline{\includegraphics[width=1\linewidth]{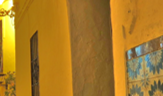}}
\end{minipage}
\begin{minipage}[b]{0.16\linewidth}
  \centering
  \centerline{\includegraphics[width=1\linewidth]{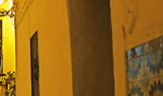}}
\end{minipage}\\
\begin{minipage}[b]{0.16\linewidth}
  \centering
  \centerline{\includegraphics[width=1\linewidth]{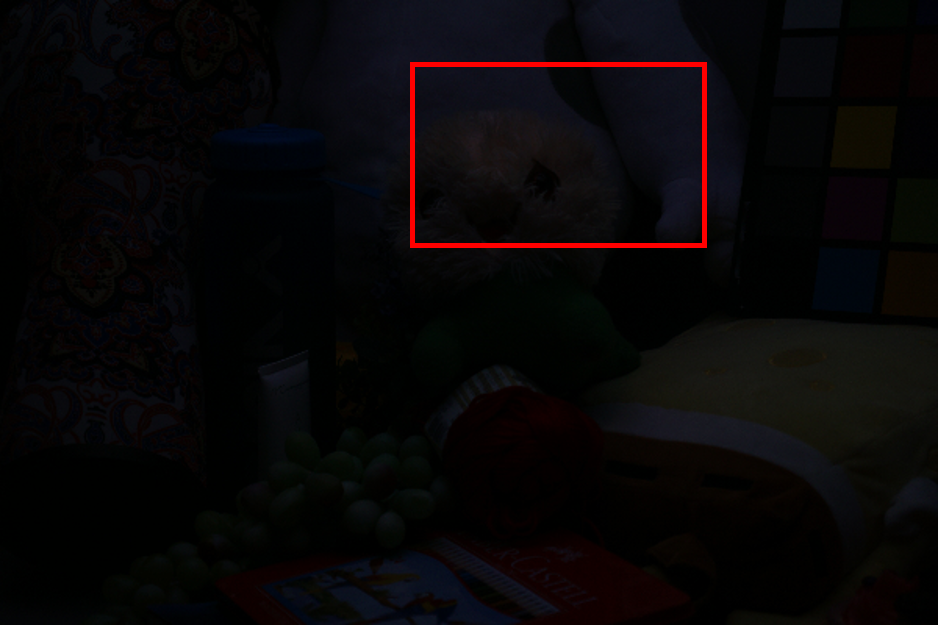}}
\end{minipage}
\begin{minipage}[b]{0.16\linewidth}
  \centering
  \centerline{\includegraphics[width=1\linewidth]{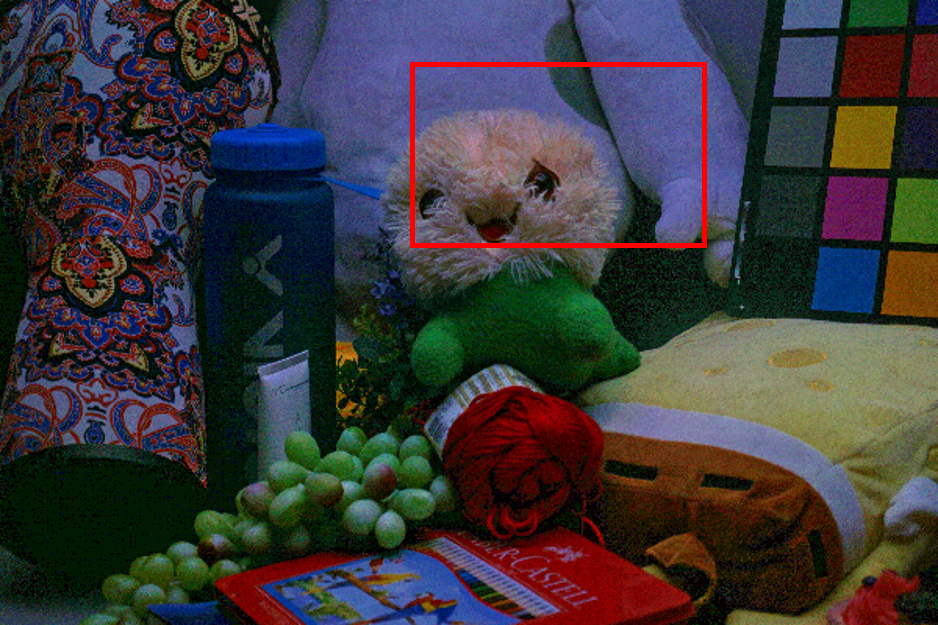}}
\end{minipage}
\begin{minipage}[b]{0.16\linewidth}
  \centering
  \centerline{\includegraphics[width=1\linewidth]{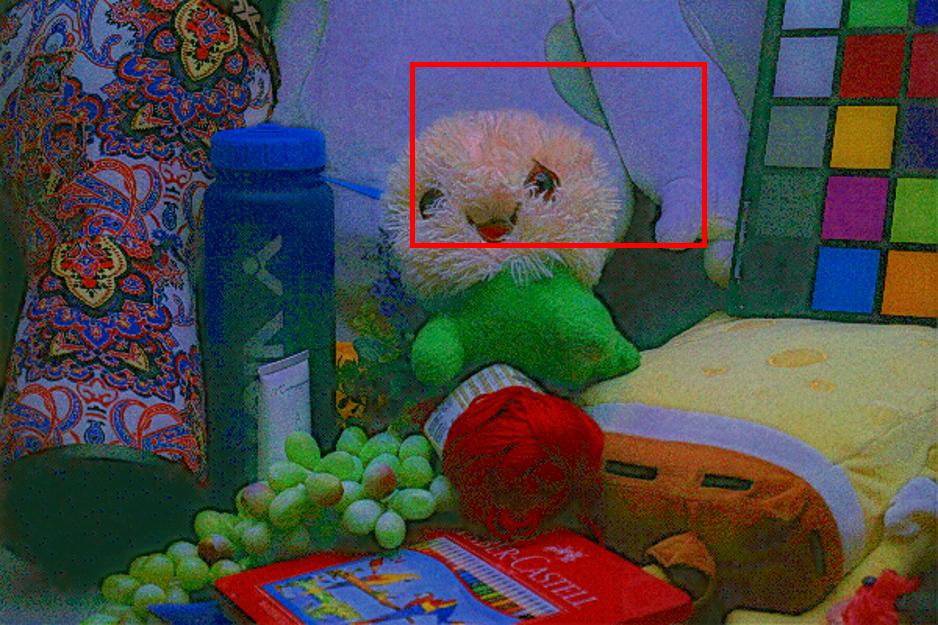}}
\end{minipage}
\begin{minipage}[b]{0.16\linewidth}
  \centering
  \centerline{\includegraphics[width=1\linewidth]{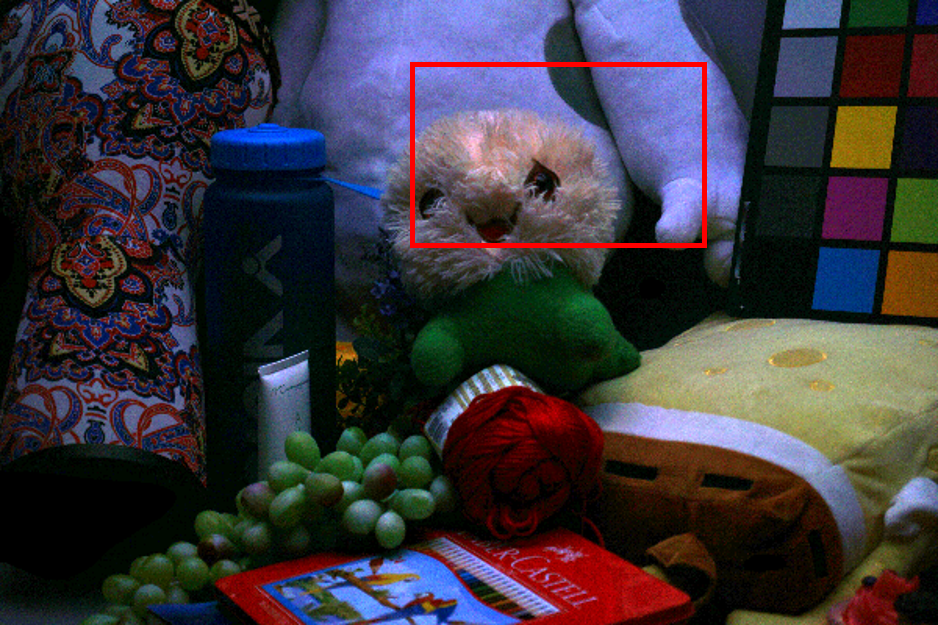}}
\end{minipage}
\begin{minipage}[b]{0.16\linewidth}
  \centering
  \centerline{\includegraphics[width=1\linewidth]{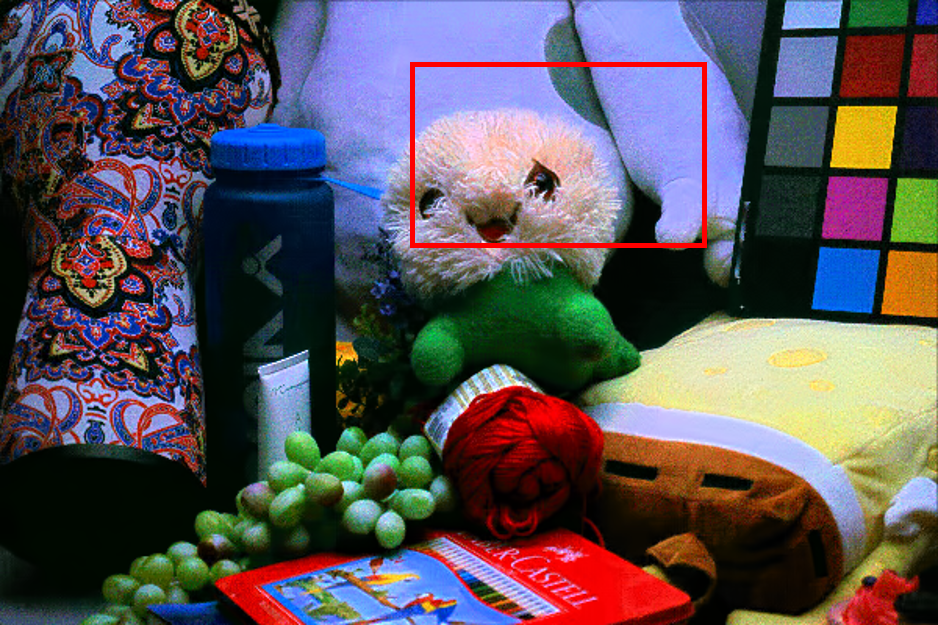}}
\end{minipage}\\
\begin{minipage}[b]{0.16\linewidth}
  \centering
  \centerline{\includegraphics[width=1\linewidth]{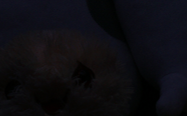}}
  \centerline{(a) Input}\medskip
\end{minipage}
\begin{minipage}[b]{0.16\linewidth}
  \centering
  \centerline{\includegraphics[width=1\linewidth]{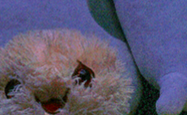}}
  \centerline{(b) PBS}\medskip
\end{minipage}
\begin{minipage}[b]{0.16\linewidth}
  \centering
  \centerline{\includegraphics[width=1\linewidth]{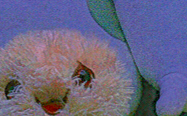}}
  \centerline{(c) Retinex-Net}\medskip
\end{minipage}
\begin{minipage}[b]{0.16\linewidth}
  \centering
  \centerline{\includegraphics[width=1\linewidth]{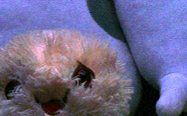}}
  \centerline{(d) ExCNet}\medskip
\end{minipage}
\begin{minipage}[b]{0.16\linewidth}
  \centering
  \centerline{\includegraphics[width=1\linewidth]{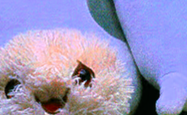}}
  \centerline{(e) SID-NISM}\medskip
\end{minipage}
\caption{Compared with the results obtained by PBS \cite{zhang2018high}, Retinex-Net \cite{wei2018deep} and ExCNet \cite{zhang2019zero} on two natural images: \textit{street} from LIME dataset and \textit{toy} from LOL dataset. Below are the enlarged detailed images in the red box. (Best viewed on screen)}
\label{fig:vision2}
\end{figure*}

In order to facilitate readers to visually compare the results of different low-light image enhancement approaches, several examples are exhibited in Fig.~\ref{fig:vision1} and ~\ref{fig:vision2}.

In general, it is clear that SID-NISM could brighten up the dark regions of low-light images successfully. More importantly, it can also restore the strong contrast of scene content sufficiently. For example, taking \textit{candle} in Fig.~\ref{fig:vision1} as input, the contrast between the candle in the foreground and the glass cup in the background is clearer and stronger in the result of SID-NISM compared with other methods. Another example is the \textit{toy} in Fig.~\ref{fig:vision2}. It is obvious that the contrast between the toy's face and the background is stronger in our result. This property renders the results of our method seem more active and attractive, which is consistent with the user study about the image overall perceptions.

In addition, as discussed in Sect.~\ref{sect:userstudy}, SID-NISM could restore the overall quality of low-light images on the premise of introducing less unexpected artifacts, including preserving clear boundary (glass cup in \textit{candle}), avoiding gray shadow (the inside of the door in \textit{street}) and halo (boundary of the door in \textit{street}). Especially, our results contain less noise such as the sky in \textit{street} and the white background in \textit{toy}. Combining with the above intuitive observation and data statistics in the user study, it can be seen that our method is more robust with less artifacts, which makes our framework more competitive and practical in both people's daily use and commercial cases.

\subsection{Objective Evaluation Indexes}
\begin{table*}[]
    \centering
    \begin{tabular}{c|ccccccc}
        \toprule
         & GE & CE & GMI & GMG & NIQE & PSNR & SSIM \\
        \hline
        SRIE                & 6.5584       & 18.4951       & 56.0559                & \textcolor{red}{\textbf{6.2808}}             & 3.8777   & 12.8554  & 0.5298   \\
        LIME                & 7.2652       & 19.8994       & 93.2529                & 11.4664            & 4.2036   & \textcolor{blue}{\textbf{17.1717}}  & 0.6355   \\
        JED                 & 6.5296       & 19.0923       & 73.9586                & 4.6584             & \textcolor{red}{\textbf{3.3546}}   & 14.9944  & 0.6325   \\
        PBS                 & 7.0828       & 20.2996       & 84.7626                & 10.8144            & 4.4976   & 16.3765  & \textcolor{blue}{\textbf{0.6375}}   \\
        Retinex-Net         & 6.9750       & \textcolor{red}{\textbf{21.0380}}        & \textcolor{red}{\textbf{110.2008}}               & 12.8826            & 5.7127   & 14.4061  & 0.4932   \\
        ExCNet              & 7.0548       & 18.7087       & 87.4292                & 10.2022            & 4.4485   & 14.9694  & 0.601    \\
        SID-NISM w/o $\mathcal{L}_{R}$ and $\mathcal{L}_{N}$         & \textcolor{red}{\textbf{7.1978}}      & 18.8489       & 99.9323                & 11.1822            & 4.4493   & 17.0313  & 0.5926   \\
        SID-NISM            & \textcolor{blue}{\textbf{7.2499}}       & \textcolor{blue}{\textbf{20.7416}}       & \textcolor{blue}{\textbf{103.3472}}  & \textcolor{blue}{\textbf{9.9731}}             & \textcolor{blue}{\textbf{3.8579}}   & \textcolor{red}{\textbf{17.7576}}  & \textcolor{red}{\textbf{0.6633}}   \\
        \hline
        Reference           & 7.1949       & 21.0804       & 110.8416               & 7.1368             & 3.5593   & /        & 1        \\
        \toprule
    \end{tabular}
    \caption{Quantitative comparison on LOL and MEF datasets in terms of different metrics. GE, CE, GMI, GMG should be close to the reference images. Lower NIQE, higher PSNR and SSIM indicate better image quality. Top2 results are highlighted in bold (\textcolor{red}{Top1}-\textcolor{blue}{Top2}).}
    \label{tab:objEva}
\end{table*}

\begin{table*}[t]
    \centering
    % \footnotesize
    \begin{tabular}{c|p{1.0cm}<{\centering}p{0.8cm}<{\centering}p{0.8cm}<{\centering}p{0.8cm}<{\centering}p{0.8cm}<{\centering}p{0.8cm}<{\centering}p{0.8cm}<{\centering}p{0.8cm}<{\centering}p{0.8cm}<{\centering}p{1.0cm}<{\centering}|c}
        \toprule
         & bicycle & boat & bottle & bus & car & cat & chair & dog & motor & person & mAP \\
        \hline
        Original & 42.36 & 8.34 & 21.38 & 42.31 & 30.74 & 20.44 & 34.77 & 25.53 & 33.12 & 36.66 & 31.54\\
        SRIE & 44.75 & 10.76 & 27.26 & 54.01 & 30.80 & 21.28 & 36.96 & 26.27 & 34.02 & 38.33 & 33.92 \\
        LIME & 47.56 & 9.25 & 24.26 & 53.24 & 32.72 & 20.40 & 32.25 & 28.29 & 32.50 & 39.98 & \textbf{33.98} \\
        JED & 41.50 & 8.53 & 21.42 & 46.53 & 23.72 & 20.46 & 31.36 & 30.24 & 31.77 & 33.83 & 29.28 \\
        PBS & 47.33 & 8.31 & 23.85 & 50.43 & 33.12 & 17.65 & 33.44 & 26.81 & 33.19 & 38.29 & 33.20 \\
        Retinex-Net & 34.59 & 5.70 & 17.80 & 38.11 & 20.99 & 16.18 & 27.57 & 6.14 & 15.53 & 27.90 & 23.15 \\
        ExCNet & 39.24 & 8.86 & 19.46 & 55.37 & 32.99 & 20.68 & 32.00 & 30.13 & 32.49 & 39.04 & 32.84 \\
        SID-NISM & 47.64 & 9.69 & 18.26 & 51.36 & 32.02 & 17.19 & 36.48 & 28.53 & 32.75 & 39.58 & \textbf{33.39} \\
        \toprule
    \end{tabular}
    \caption{The performance (mAP: \%) of object detection on real low-light images and their enhanced versions by low-light image enhancement methods.}
    \label{tab:objectmap}
\end{table*}

In addition to the above subjective experiments, there are many indexes with or without reference images that can be used to evaluate the quality of the enhanced images objectively. In this section, we take use of gray entropy (GE), color entropy (CE, sum of the entropy of RGB channels), gray mean illumination (GMI), gray mean gradient (GMG), NIQE~\cite{mittal2012making}, PSNR, and SSIM~\cite{wang2004image} to compare different low-light image enhancement methods. Generally, the first four indexes could reflect the gap between the enhanced results and the reference high-light images from four aspects including the information in gray and color channels, the illumination degree and the image gradient magnitude. NIQE assesses the overall naturalness of the enhanced results and a lower value roughly corresponds to a higher overall naturalness. PSNR could verify the performance of denoising algorithms and a higher value usually indicates a better image quality. SSIM measures the structure similarity between the enhanced results and the reference images, which is in the range of [0, 1]. It should be noted that these indexes can only reﬂect the image quality in specific aspects, which may be not completely consistent with the evaluation results given by the human visual system.

The quantitative results over LOL and MEF datasets (the two datasets with reference normal-light images) are reported in Table \ref{tab:objEva}. Obviously, the proposed SID-NISM could achieve better performance compared with existing state-of-the-art methods in terms of the image information, the overall quality and the noise condition. It is worth mentioning that we also compare the performance of our framework with and without  $\mathcal{L}_{R}$ and $\mathcal{L}_{N}$ to quantitatively evaluate the merit brought by the two novel reflectance and noise terms in the loss function of SID-Net. It can be seen that they not only help improve the image color and gradient information of the enhanced results greatly, but also decrease the noise influence and promote the overall image quality.

\section{Object Detection Test}
We also take a step further to look into the potential of low-light image enhancement in improving the performance of one popular computer vision task, object detection. To do so, tests were performed on YOLOv2 platform on real low-light images and their enhanced versions by our proposed framework and other state-of-the-art methods. Here we directly used the pre-trained VOC2007+2012 model without any re-training or fine-tuning as we would like to observe the outcome of the originally optimized model when given low-light and enhanced images. Experiments were conducted on Exclusively Dark Image Dataset (ExDark), which is consist of low-light images taken in 10 different low-light environments with corresponding 12 image classes and object level annotations.

Table \ref{tab:objectmap} exhibits the mean of Average Precision (mAP) over 10 classes when detecting different versions of images. Obviously, compared to detecting objects on the original low-light images directly, the performance of the object detection model would be improved by preprocessing the inputs through our proposed low-light image enhancement method from 31.54\% to 33.39\% in terms of mAP. Among existing methods, LIME reaches the best mAP (33.98\%) while SID-NISM shows its competitive power (within 1\%) in improving the performance of downstream tasks. This test illustrates the feasibility of incorporating image enhancement as a support for practical applications.

\section{Conclusion}
In this paper, a self-supervised low-light image enhancement method called SID-NISM is proposed, which can first decompose given input images in an unsupervised way without relying on any external examples or prior training, and then enhance decomposed illumination map by a well-designed nonlinear illumination mapping function. The proposed scheme is concise yet powerful. Experimental results show that, even though our method does not need any support data or prior knowledge, it can produce visually pleasing enhancement results with less unexpected artifacts as well as a good representation of image decomposition.

{\small
\bibliographystyle{ieee_fullname}
\bibliography{egbib}
}

\end{document}